\documentclass[Afour,sageH,times]{sagej}

\usepackage{url}
\usepackage{cite}
\usepackage{graphicx}
\usepackage{amsmath}
\usepackage{arydshln}
\usepackage{listings}
\usepackage{xcolor}
\usepackage{mathtools}
\usepackage[normalem]{ulem}
\usepackage{float}
\usepackage{bbm}
\usepackage{amssymb}
\usepackage{caption}
\usepackage{subcaption}
\usepackage{moreverb,url}

\usepackage[colorlinks,bookmarksopen,bookmarksnumbered,citecolor=red,urlcolor=red]{hyperref}

\usepackage{comment}
\usepackage[english]{babel}
\usepackage{amsthm}
\usepackage[linesnumbered,ruled,vlined,titlenumbered,longend, procnumbered]{algorithm2e}
\definecolor{LB}{RGB}{65,105,225}
\definecolor{LG}{rgb}{0.13, 0.55, 0.13}
\definecolor{LB}{RGB}{65,105,225}
\definecolor{LO}{RGB}{210,127,0}
\definecolor{LP}{cmyk}{0, 0.7808, 0.4429, 0.1412}
\definecolor{LG}{rgb}{0.13, 0.55, 0.13}
\SetKw{Exit}{exit}
\SetKw{Break}{break}
\newtheorem{defn}{Definition}
\newtheorem{assumption}{Assumption}
\newtheorem{rmrk}{Remark}

\newtheorem{exmp}{Example}
\newtheorem{thm}{Theorem}
\newtheorem{lemma}{Lemma}

\newtheorem{prob}{Problem}
\SetKw{KwBy}{by}
\SetKw{KwTo}{to}

\begin{document}
\raggedbottom
\runninghead{Sewlia et al.}

\title{\LARGE \bf 
MAPS$^\mathbf{2}$:  Multi-Robot Autonomous Motion Planning under Signal Temporal Logic Specifications
}

\author{Mayank Sewlia\affilnum{1}, Christos K. Verginis\affilnum{2} and Dimos V. Dimarogonas\affilnum{1}}

\affiliation{\affilnum{1}Division of Decision and Control, School of EECS, KTH Royal Institute of Technology, Stockholm, Sweden. \\
\affilnum{2}Division of Signals and Systems, Department of Electrical Engineering, Uppsala University,
Uppsala, Sweden.\\
\textbf{Acknowledgements:} This work was supported by the ERC CoG LEAFHOUND, the Swedish Research Council (VR), the Knut och Alice Wallenberg Foundation (KAW)
and the H2020 European Project CANOPIES.}
\corrauth{Mayank Sewlia, Division of Decision and Control, 
School of EECS, 
KTH Royal Institute of Technology, 
Stockholm, Sweden.}

\email{sewlia@kth.se}

%\thispagestyle{empty}
%\pagestyle{empty}

%%%%%%%%%%%%%%%%%%%%%%%%%%%%%%%%%%%%%%%%%%%%%%%%%%%%%%%%%%%%%%%%%%%%%%%%%%%%%%%%
\begin{abstract}
This article presents MAPS$^2$: a distributed algorithm that allows multi-robot systems to deliver coupled tasks expressed as Signal Temporal Logic (STL) constraints. Classical control theoretical tools addressing STL constraints either adopt a limited fragment of the STL formula or require approximations of min/max operators. Meanwhile, works maximising robustness through optimisation-based methods often suffer from local minima, thus relaxing any completeness arguments due to the NP-hard nature of the problem. Endowed with probabilistic guarantees, MAPS$^2$ provides an autonomous algorithm that iteratively improves the robots' trajectories. The algorithm selectively imposes spatial constraints by taking advantage of the temporal properties of the STL. The algorithm is distributed in the sense that each robot calculates its trajectory by communicating only with its immediate neighbours as defined via a communication graph. We illustrate the efficiency of MAPS$^2$ by conducting extensive simulation and experimental studies, verifying the generation of STL satisfying trajectories.
\end{abstract}

\maketitle
%%%%%%%%%%%%%%%%%%%%%%%%%%%%%%%%%%%%%%%%%%%%%%%%%%%%%%%%%%%%%%%%%%%%%%%%%%%%%%%%
\section{Introduction}

Autonomous robots can solve significant problems when provided with a set of guidelines. These guidelines can be derived from either the physical constraints of the robot, such as joint limits, or imposed as human-specified requirements, such as pick-and-place objects. An efficient method of imposing such guidelines is by using logic-based tools, which enable reasoning about the desired behaviour of robots. These tools help us describe the behaviour of a robot at various levels of abstraction, such as interactions between its internal components to the overall high-level behaviour of the robot \cite{lamport1983what}. This strong expressivity helps us efficiently encode complex mission specifications into a logical formula. Recent research has focused on utilising these logic-based tools to express requirements on the behaviour of robots. Once these requirements are established, algorithms are developed to generate satisfying trajectories. Such is the focus of our work. 

%In the context of autonomous robots, the behaviour of the robot is to reason about its high-level trajectories through a process known as planning. 
%Utilising logic-based tools, the trajectory can be described in discrete or continuous time, in discrete or continuous space, or in relation to events in time. 

Examples of logic-based tools include formal languages, such as Linear Temporal Logic (LTL), Metric Interval Temporal Logic (MITL), and Signal Temporal Logic (STL). The main distinguishing feature between these logics is their ability to encode time. While LTL operates in discrete-time and discrete-space domain, MITL operates in the continuous-time domain but only enforces qualitative space constraints. On the other hand, STL allows for the expression of both qualitative and quantitative semantics of the system in both continuous-time and continuous-space domains \cite{maler2004monitoring}.  STL thus provides a natural and compact way to reason about a robot's motion since it operates in a continuously evolving space-time environment. Additionally, STL is accompanied by a robustness metric which allows us to determine the extent of satisfaction compared to only absolute satisfaction \cite{donze2013signal}. 

\begin{figure}
    \includegraphics[width=\columnwidth]{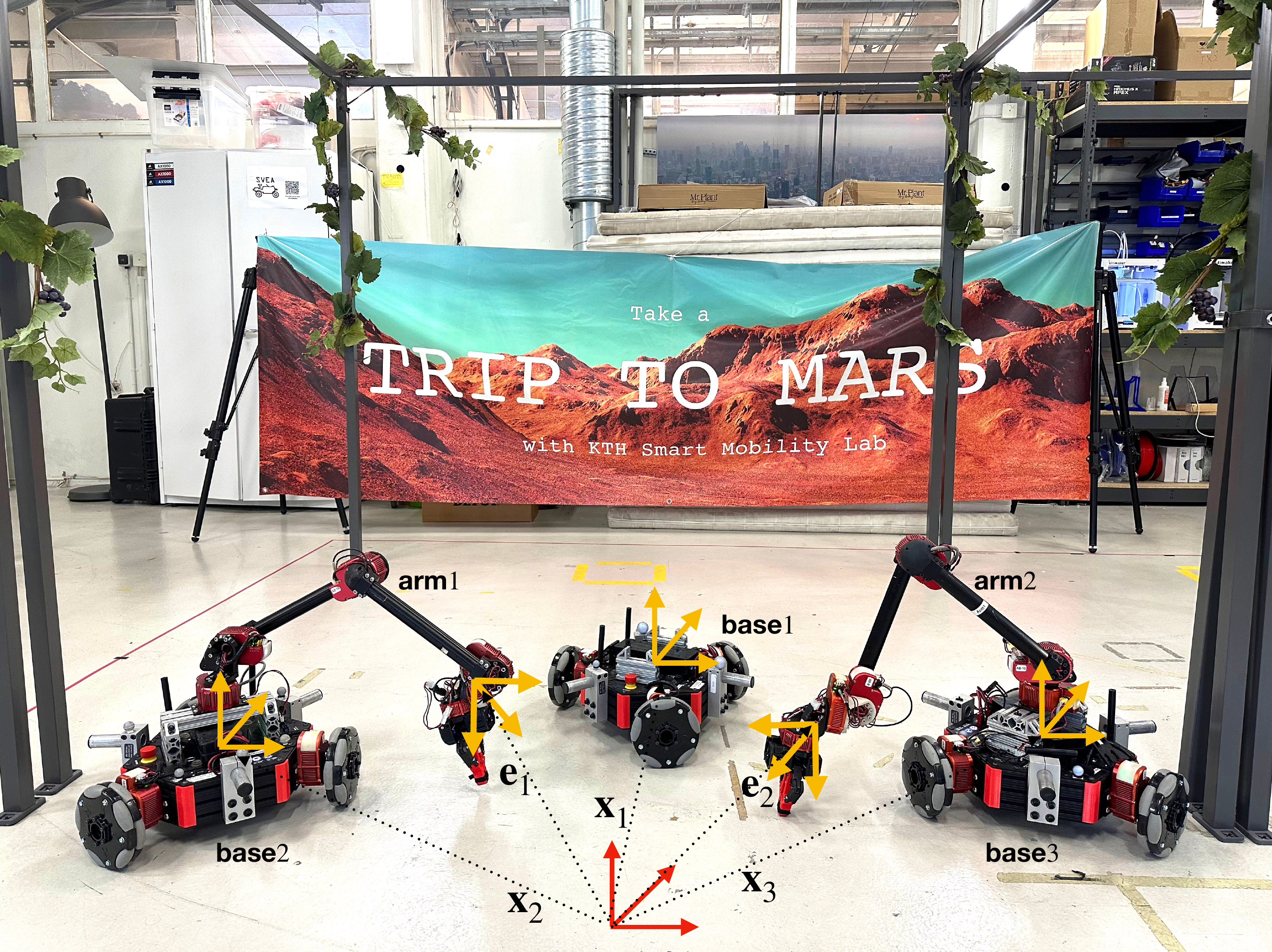}
    \caption{Experimental setup with three mobile bases and two 6-dof manipulators.}
    \label{fig:setup}
\end{figure}

Another important property of autonomous robots is their ability to coordinate and work in teams. The use of multiple robots is often necessary in situations where a single robot is either insufficient, the task is high-energy demanding, or unable to physically perform certain tasks. 
However, multi-robot systems present their own set of challenges, such as communication overload, the need for a central authority for commands, and high computational demands. The challenge, therefore, is to derive solutions for multi-robot problems utilising logic-based tools, ensuring the achievement of specified high-level behaviour. 

In this article, we propose \textbf{MAPS}$^2$ - \textbf{M}ulti-{R}obot \textbf{A}utonomous Motion \textbf{P}lanning under \textbf{S}ignal Temporal Logic \textbf{S}pecifications - to address the multi-robot motion planning problem subject to coupled STL contraints. The algorithm encodes these constraints into an optimisation function and selectively activates them based on the temporal requirements of the STL formula. While doing so, each robot only communicates with its neighbours and iteratively searches for STL satisfying trajectories. 
The algorithm ensures distributed trajectory generation to satisfy STL formulas that consist of coupled constraints for multiple robots. The article's contributions are summarised in the following attributes:
{
\begin{itemize}
\item {The algorithm's effectiveness lies in its ability to distribute STL planning for multiple robots and in providing a mechanism to decouple the STL formula among robots, thereby facilitating the distribution of tasks.}
\item {As opposed to previous work, it covers the entire STL formula and is not limited to a smaller fragment. It reduces conservatism by eliminating the need for approximations of max/min operators and samples in continuous time to avoid abstractions. }
\item{It incorporates a wide range of coupled constraints (both linear and nonlinear) into the distributed optimisation framework, enabling the handling of diverse tasks such as pick-and-place operations and time-varying activities like trajectory tracking.}
\item {We present extensive simulation and hardware experiments that demonstrate the execution of complex tasks using MAPS$^2$.}
\end{itemize}
}
Additionally, the algorithm presented is sound, meaning that it produces a trajectory that meets the STL formula and is probabilistically complete, meaning that it will find such a trajectory if one exists. 

In our prior study \cite{10156470}, we addressed the STL motion planning problem for two coupled agents. There, we extended the conventional Rapidly-exploring Random Trees (RRT) algorithm to sample in both the time and space domains. Our approach incrementally built spatio-temporal trees through which we enforced space and time constraints as specified by the STL formula. The algorithm employed a sequential planning method, wherein each agent communicated and waited for the other agent to build its tree. In contrast, the present work addresses the STL motion planning problem for multiple robots. Here, our algorithm adopts a distributed optimisation-based approach, where spatial and temporal aspects are decoupled to satisfy the STL formula. Instead of constructing an incremental tree, as done in the previous work, we introduce a novel metric called the \emph{validity domain} and initialise the process with an initial trajectory. In the current research, we only incorporate the STL parse tree and the Satisfaction variable tree from our previous work (Section \ref{subsec:stl_parse} here). Additionally, we present experimental validation results and introduce a novel STL verification architecture.

The rest of the paper is organised as follows. Section \ref{sec:related_work} presents the related work. Section \ref{sec:prelims} presents preliminaries and formulates the problem of this work, Section \ref{sec:decompose} presents how we decompose the STL formula into temporal and spatial constraints, Section \ref{sec:main} presents the main algorithm MAPS$^2$ along with analyses of the algorithm, Section \ref{sec:simulations} presents simulations of various tasks encountered in a robotics problem, while Section \ref{sec:experiments} presents the experimental validation on a real multi-robot setup. Finally, Section \ref{sec:conclusion} concludes the paper. 

\section{Related Work}\label{sec:related_work}

In the domain of single-agent motion planning, different algorithms have been proposed to generate safe paths for robots. Sampling-based algorithms, such as CBF-RRT \cite{yang2019sampling}, have achieved success in providing a solution to the motion planning problem in dynamic environments. However, they do not consider high-level complex mission specifications. Works that impose high-level specifications in the form of LTL, such as \cite{6697120, 5509503, 6697051, FAINEKOS2009343}, resort to a hybrid hierarchical control regime resulting in abstraction and explosion of state-space. While a mixed integer program can encode this problem for linear systems and linear predicates \cite{6907641}, the resulting algorithm has exponential complexity, making it impractical for high-dimensional systems, complex specifications, and long duration tasks. To address this issue, \cite{kurtz2022mixed} proposes a more efficient encoding for STL to reduce the exponential complexity in binary variables. Additionally, \cite{lindemann2017robust} introduces a new metric, discrete average space robustness, and composes a Model Predictive Control (MPC) cost function for a subset of STL formulas.

In multi-agent temporal logic control, works such as \cite{verginis2018timed, 5238617} employ workspace discretisation and abstraction techniques, which we avoid in this article due to it being computationally demanding. Some approaches to STL synthesis involve using mixed-integer linear programming (MILP) to encode constraints, as previously explored in \cite{belta2018, raman2014model, 7447084}. However, MILPs are computationally intractable when dealing with complex specifications or long-term plans because of the large number of binary variables required in the encoding process.   The work in \cite{sun2022multi} encodes a new specification called multi-agent STL (MA-STL) using mixed integer linear programs (MILP). However, the predicates here depend only on the states of a single agent, can only represent polytope regions, and finally, temporal operations can only be applied to a single agent at a time. In contrast, this work explores coupled constraints between robots and predicates are allowed to be of nonlinear nature.

As a result, researchers have turned to transient control-based approaches such as gradient-based, neural network-based, and control barrier-based methods to provide algorithms to tackle the multi-robot STL satisfaction problem \cite{kurtz2022mixed}. Such approaches, at the cost of imposing dynamical constraints on the optimisation problem, often resort to using smooth approximations of temporal operators at the expense of completeness arguments or end-up considering only a smaller fragment of the syntax \cite{8264095, 9655231, 9838497, 8431695}. STL's robust semantics are used to construct cost functions to convert a synthesis problem to an optimisation problem that benefits from gradient-based solutions. However, such approaches result in non-smooth and non-convex problems and solutions are prone to  local minima \cite{gilpin2020smooth}. In this work, we avoid approximations and consider the full expression of the STL syntax. The proposed solution adopts a purely geometrical approach to the multi-robot STL planning problem. Our current focus is directed towards the planning problem, specifically the generation of trajectories that fulfil STL constraints, rather than the dynamical constraints or the precise control techniques used to execute the trajectory.\\

%%%%%%%%%%%%%%%%%%%%%%%%%%%%%%%%%%%%%%%%%%%%%%%%%%%%%%%%%%%%%%%%%%%%%%%%%%%%%%%%

%%%%%%%%%%%%%%%%%%%%%%%%%%%%%%%%%%%%%%%%%%%%%%%%%%%%%%%%%%%%%%%%%%%%%%%%%%%%%%%%
\noindent\textbf{Notations:}
The set of natural numbers is denoted by $\mathbb{N}$ and the set of real numbers by $\mathbb{R}$. With $n\in\mathbb{N}$, $\mathbb{R}^n$ is the set of $n$-coordinate real-valued vectors and $\mathbb{R}_{+}^n$ is the set of real $n$-vector with non-negative elements. {The cardinality of a set $A$ is denoted by $|A|$}. If $a\in\mathbb{R}$ and $[b,c]\in\mathbb{R}^{2}$, the Kronecker sum is defined as $a\oplus[b,c]=[a+b, a+c]\in \mathbb{R}^{2}$. We further define the Boolean set as $\mathbb{B}=\{\top,\bot\}$ (True, False). The acronym \textit{DOF} stands for degrees of freedom.  %By $\mathcal{W}_i\in \mathbb{R}^{n}$ denote the workspace of each robot, i.e. the space in which planning takes place, and let $\mathcal{W}= \mathcal{W}_i$.

%%%%%%%%%%%%%%%%%%%%%%%%%%%%%%%%%%%%%%%%%%%%%%%%%%%%%%%%%%%%%%%%%%%%%%%%%%%%%%%%

\section{Preliminaries and Problem Formulation}\label{sec:prelims}

In this section, we start by introducing STL and STL parse tree, followed by the problem formulation.

\subsection{Signal Temporal Logic (STL)}\label{subsec:stl}

Let $\mathbf{x}:\mathbb{R}_{+}\to\mathbb{R}^n$ be a continuous-time signal.
Signal temporal logic \cite{maler2004monitoring} is a predicate-based logic with the following syntax:

\begin{equation}\label{eq:stl_general}
    \varphi = \top \ |\ \mu^h\ |\ \neg \varphi\ |\  \varphi_1\mathcal{U}_{[a,b]}\varphi_2\ |\ \varphi_1\land\varphi_2\
\end{equation}
where $\varphi_1,\ \varphi_2$ are STL formulas and $\mathcal{U}_{[a,b]}$ encodes the operator \textit{until}, with $0 \leq a < b < \infty$; 
$\mu^h$ is a predicate of the form $\mu^h:\mathbb{R}^n\to \mathbb{B}$ defined by means of a vector-valued predicate function $h:\mathbb{R}^n\to \mathbb{R}$ as

\begin{equation} \label{eq:mu}
    \mu^h=\begin{cases}\top & h(\mathbf{x}(t))\leq 0\\ \bot & h(\mathbf{x}(t))>0\end{cases}.
\end{equation}  
{ The satisfaction relation $(\mathbf{x},t)\models \varphi$ indicates that signal $\mathbf{x}$ satisfies $\varphi$ at time $t$ and is defined recursively as follows:

\begin{alignat*}{2}
    &(\mathbf{x},t)\models \mu^h &&\Leftrightarrow  h(\mathbf{x}(t))\leq 0\\
    &(\mathbf{x},t)\models \neg \varphi &&\Leftrightarrow  \neg((\mathbf{x},t)\models \varphi)\\
    &(\mathbf{x},t)\models \varphi_1 \land \varphi_2  &&\Leftrightarrow (\mathbf{x},t)\models \varphi_1 \land (\mathbf{x},t)\models \varphi_2\\
    & (\mathbf{x},t)\models \varphi_1 \mathcal{U}_{[a,b]}\varphi_2 &&\Leftrightarrow \exists t_1\in[t+a, t+b] \text{ s.t. } (\mathbf{x},t_1)\models \varphi_2 \\ &&& \quad \land \forall t_2\in[t,t_1], (\mathbf{x},t_2)\models \varphi_1. 
\end{alignat*}
}

We also define the operators \textit{disjunction}, \textit{eventually}, and \textit{always}  as $\varphi_1 \lor \varphi_2 \equiv \neg(\neg \varphi_1 \land \neg \varphi_2)$, $\mathcal{F}_{[a,b]}\varphi \equiv \top \mathcal{U}_{[a,b]} \varphi$,  and $\mathcal{G}_{[a,b]}\varphi \equiv \neg \mathcal{F}_{[a,b]}\neg \varphi$, respectively. Each STL formula is valid over a time horizon defined as follows.

\begin{defn}[ \cite{madsen2018metrics} ]\label{defn:time_horizon}
The time horizon $\mathrm{th}(\varphi)$ of an STL formula $\varphi$ is recursively defined as,
\begin{equation}
    \mathrm{th}(\varphi) = 
    \begin{cases} 0,& \text{if}\ \varphi=\mu\\ 
    \mathrm{th}(\varphi_1),& \text{if}\  \varphi=\neg\varphi_1\\ 
    \max\{\mathrm{th}(\varphi_1), \mathrm{th}(\varphi_2)\},& \text{if}\  \varphi=\varphi_1\land\varphi_2\\ %b+\mathrm{th}(\varphi_1), & \text{if}\  \varphi\in\{\mathcal{G}_{[a,b]}\varphi_1,\\ &\quad \quad \quad \mathcal{F}_{[a,b]}\varphi_1 \}.\\
    b+\max\{\mathrm{th}(\varphi_1),\mathrm{th}(\varphi_2)\},& \text{if}\ \varphi = \varphi_1\mathcal{U}_{[a,b]}\varphi_2.
    \end{cases}
\end{equation}
\end{defn}
\noindent In this work, we consider only time bounded temporal operators,  i.e., when $\mathrm{th}(\varphi)<\infty$. In the case of unbounded STL formulas, it is only possible to either falsify an \emph{always} operator or satisfy an \emph{eventually} operator in finite time, thus we consider only bounded time operators. Next, we state a common assumption regarding the STL formula.
\begin{assumption}\label{ass:pnf}
The STL formula is in \emph{positive normal form} i.e., it does not contain the negation operator. 
\end{assumption}
The above assumption does not cause any loss of expression of the STL syntax \eqref{eq:stl_general}. As shown in \cite{7447084}, any STL formula can be written in positive normal form by moving the negation operator to the predicate.

%%%%%%%%%%%%%%%%%%%%%%%%%%%%%%%%%%%%%%%%%%%%%%%%%%%%%%%%%%%%%%%%%%%%%%%%%%%%%%%%

\subsection{STL Parse Tree}\label{subsec:stl_parse}
An STL parse tree is a tree representation of an STL formula \cite{10156470}. It can be constructed as follows:

\begin{itemize}
    \item Each node is either a temporal operator node $\{\mathcal{G}_I, \mathcal{F}_I\}$, a logical operator node $\{\lor, \land, \neg\}$, or a predicate node $\{\mu^h\}$, where $I\subset \mathbb{R}$ is a closed interval;
    \item temporal and logical operator nodes are called \textit{set} nodes;
    \item a root node has no parent node and a leaf node has no child node. The leaf nodes constitute the predicate nodes of the tree.
\end{itemize}

A path in a tree is a sequence of nodes that starts at a root node and ends at a leaf node. The set of all such paths constitutes the entire tree. A subpath is a path that starts at a set node and ends at a leaf node; a subpath could also be a path. The resulting formula from a subpath is called a subformula of the original formula. In the following, we denote any subformula of an STL formula $\varphi$ by $\bar{\varphi}$. Each set node is accompanied by a satisfaction variable $\tau:\bar{\varphi}\to\{+1,-1\}$ and each leaf node is accompanied by a predicate variable $\pi=\mu^h$ where $h$ is the corresponding predicate function. A signal $\mathbf{x}$ satisfies a subformula $\bar{\varphi}$ if $\tau=+1$ corresponding to the set node where the subpath of $\bar{\varphi}$ begins. Similarly, 
$\tau(\text{\emph{root($\varphi$)}})=+1 \Leftrightarrow (\mathbf{x},t)\models\varphi$ where \emph{root} is the root node of $\varphi$. An analogous tree of satisfaction and predicate variables can be drawn, called \emph{satisfaction variable tree}. {The satisfaction variable tree borrows the same tree structure as the STL parse tree. Each set node from the STL parse tree maps uniquely to a satisfaction variable $\tau_i$ and each leaf node maps uniquely to a predicate variable $\pi_i$, where $i$ is an enumeration of the nodes in the satisfaction variable tree}. An example of construction of such trees is shown below. 

\begin{figure}
     \centering
     \includegraphics[width=0.7\columnwidth]{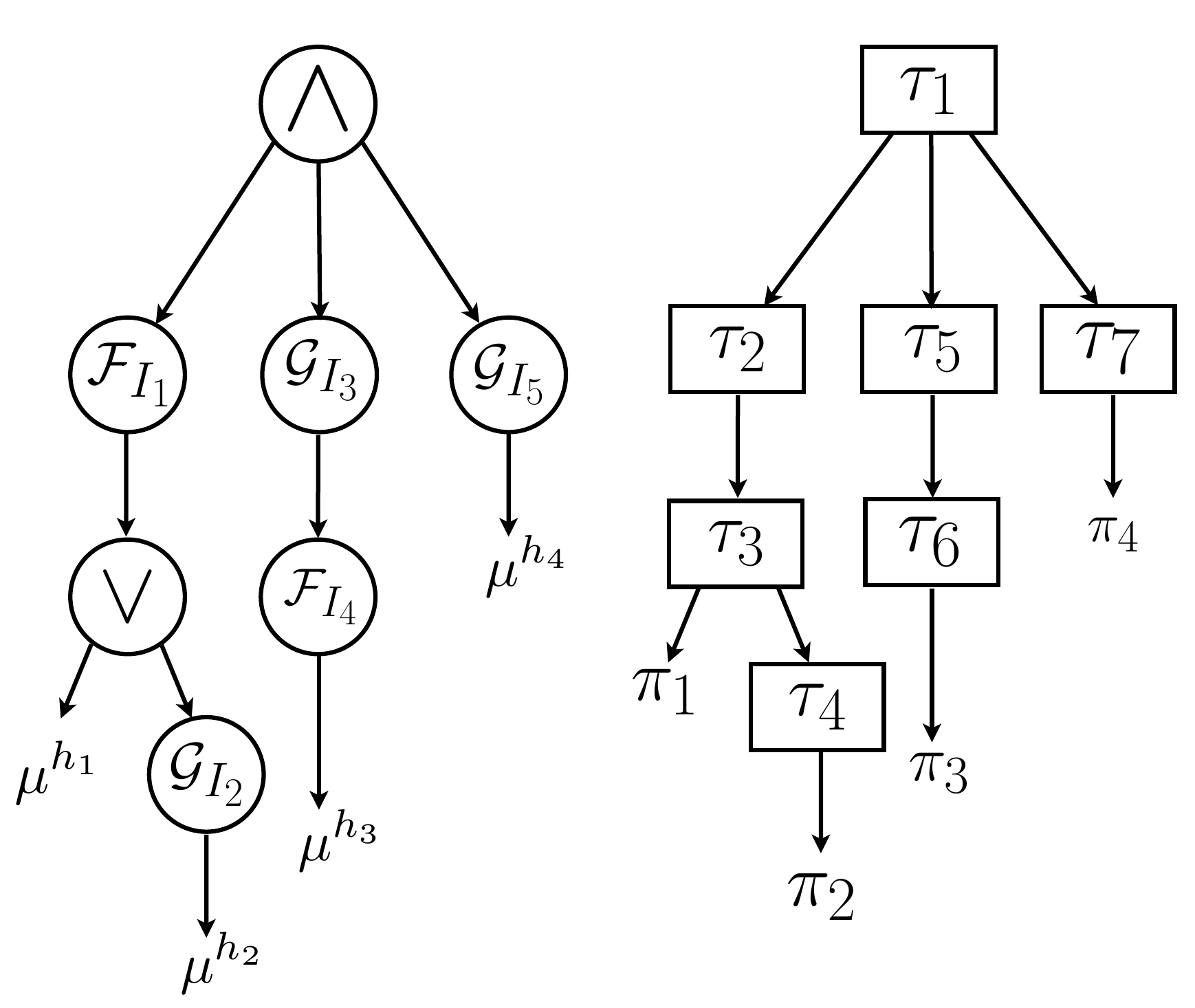}
     \caption{STL parse tree and satisfaction variable tree for the formula in \eqref{eq:stl_eg}.}
     \label{fig:parse}
\end{figure}

\begin{exmp}
The STL parse tree and the satisfaction variable tree for the STL formula 
\begin{align}\label{eq:stl_eg}
    \varphi &= \mathcal{F}_{I_1}\Big({\mu^{h_1}}\lor \mathcal{G}_{I_2} ({\mu^{h_2}})\Big) \land \mathcal{G}_{I_3}\mathcal{F}_{I_4}({\mu^{h_3}})\land \mathcal{G}_{I_5}({\mu^{h_4}}).
\end{align}
are shown in Figure \ref{fig:parse}. From the trees, one obtains the implications $\tau_2=+1\implies (\mathbf{x},t)\models \mathcal{F}_{I_1}\Big({\mu^{h_1}}\lor \mathcal{G}_{I_2} ({\mu^{h_2}})\Big)$, and  $\tau_7=+1\implies (\mathbf{x},t)\models\mathcal{G}_{I_5}({\mu^{h_4}})$.
\end{exmp}

\subsection{Problem Formulation}\label{subsec:problem}
{We consider a team of $\mathbf{N}$ robots, where each robot has state $\mathbf{x}_i\in\mathcal{W}_i\subset \mathbb{R}^{n_i}$, $i\in\{1,\dots,\mathbf{N}\}$ and $n_i$ is the number of degrees of freedom of robot $i$. The overall state vector is then $\mathbf{x}:=[\mathbf{x}_1^\top\ \cdots\ \mathbf{x}_{\mathbf{N}}^\top]^\top$ evolving in a workspace $\mathcal{W}=\mathcal{W}_1\times\cdots \times \mathcal{W}_{\mathbf{N}}$ and we denote by $n=n_1+\cdots+n_{\mathbf{N}}$ the number of degrees of freedom of the multi-robot system. We consider the STL formula of the form \eqref{eq:stl_general} with a total of $K$ predicates, 
\begin{equation}\label{state_constraints}    
    \mu^{h^{(k)}}=\begin{cases}\top & h^{(k)}(\mathbf{x}(t))\leq 0\\ \bot & h^{(k)}(\mathbf{x}(t))>0\end{cases}, \quad k=1,\dots,K.
\end{equation}
Before we present the problem statement, we will introduce the multi-robot STL notation and the communication structure of the multi-robot system. In this direction, define a support set for each predicate function $h^{(k)}(\mathbf{x})$ that captures all the robots upon which the predicate function imposes constraints. Define a projection matrix $E_i\in\mathbb{R}^{n\times n_i}$ such that $E_i^\top \mathbf{x}=\mathbf{x}_i$. The matrix $E_i$ takes the form,
\[
E_i=\begin{bmatrix}\mathbf{0}_{n_1}\\ \vdots\\ I_{n_i}\\ \vdots \\ \mathbf{0}_{n_\mathbf{N}}\end{bmatrix}\in\mathbb{R}^{n\times n_i},
\]
i.e. $E_i v = \begin{bmatrix}0,\dots,0,v,0\dots,0\end{bmatrix}^\top$ inserts any $v\in\mathbb{R}^{n_i}$ at the $i$-th position of the vector. 
The support set is then defined for each predicate function $h^{(k)}$ as,
\begin{align*}
\mathbf{S}_k := \Big\{i\in\mathcal{I}\ |\ \exists \mathbf{x}\in\mathbb{R}^n, v\in\mathbb{R}^{n_i}, \epsilon>0\\:h^{(k)}(\mathbf{x}+\epsilon E_i v)\neq h^{(k)}(\mathbf{x})\Big\}.
\end{align*}
The support set thus captures all robots $i$ for which some perturbation confined to their own state $\mathbf{x}_i$ can change the value of the predicate function $h^{(k)}$. If a predicate function $h^{(k)}$ imposes constraints on the states of multiple robots, i.e., $|\mathbf{S}_k|\geq 2$, then we say that the predicate function is coupled. \\
\begin{exmp}
For an STL formula $\varphi = (\|x_1-x_2\|>5)\ \land (\|x_2-x_3\|<2)$, $h^{(1)}=5-\|x_1-x_2\|$ and $h^{(2)}=\|x_2-x_3\|-2$. Then $\mathbf{S}_1=\{1,2\}$ and $\mathbf{S}_2=\{2,3\}$.\\
\end{exmp}
Let $K_c \leq K$ denote the number of coupled predicate functions, and let these be indexed as $h^c_j$, where $j \in \{1, \dots, K_c\}$. For each $j$, there exists an index $k_j \in \{1, \dots, K\}$ such that $h^c_j = h^{(k_j)}$; let $\mathbf{S}_{k_j} \subseteq \{1, \dots, \mathbf{N}\}$ denote the support set of $h^c_j$, i.e., the set of robots whose states appear in $h^c_j$. Each robot $i \in \mathbf{S}_{k_j}$ then uses a local copy $h^c_{i,j} := h^c_j$. This local copy is provided to the robots manually offline prior the start of the algorithm. The robots can also obtain the functions $h^d_{i,l}$ and $h^c_{i,j}$ that are coupled to their own states directly from $\varphi$. This could, for example, be done by defining a \textit{regular expression}(regex) pattern and extracting predicate functions that involve the states $\mathbf{x}_i$ \cite{goyvaerts_regex_tutorial}.}

{Next, let $L_i$ be the number of independent predicate functions that involve only the states of robot $i$, with $\sum_{i=1}^{\mathbf{N}} L_i = K-K_c$. Such predicate functions are indexed as $h^d_{i,l}$, $i\in\{1,\dots,\mathbf{N}\}$ and $l\in\{1,\dots,L_i\}$. 

Let $\Pi_{k_j}$ be the projection that keeps the indicated state components of the robots in $\mathbf{S}_{k_j}$ only, i.e. $\Pi_{{k_j}} \mathbf{x}:= \{E_m^\top \mathbf{x}\ |\ m\in\mathbf{S}_{k_j}\}$.
The predicate function constraints for each robot $i$ are then defined as follows
\begin{equation}\label{eq:labelled_constraint_predicates}
h^d_{i,l}(\mathbf{x}_i)\leq 0\quad \text{and}\quad  h^c_{i,j}(\Pi_{{k_j}}\mathbf{x})\leq 0
\end{equation}
for all $i\in\{1,\dots,\mathbf{N}\}$, $l\in\{1,\dots,L_i\}$ and $j\in\{1,\dots,K_c| i\in\mathbf{S}_{k_j}\}$. The coupled predicate functions $h^c_{i,j}$ can reflect physical interactions between the robots if the constraint is such, for example, if  $h^c_{i,j}$ specifies an obstacle avoidance constraint or an object handover task. 
\begin{exmp}\label{exmp:labelled_predicates}
    Consider the STL formula, 
    \begin{align*}
        \varphi = (\|x_1-x_2\|<1) &\land  (\|x_2-x_3\|<1)\\
                 (\|x_3-x_4\|<1) &\land  (\|x_4-x_1\|<1) \\
            (\|x_1\|<1) &\land  (\|x_2-x_3-x_4\|<1)\\
            (\|x_2\|<1) &\land  (\|x_5\|<1)
    \end{align*}
    For the above STL formula, $K=8$ and $K_c=5$. The number of independent predicates for each robot are $L_1 =1$, $L_2=1$, $L_3=L_4=0$ and $L_5=1$. 
    Table \ref{tab:constraints} depicts the labelled predicates for the above STL formula.
    \begin{table}[ht]
        \centering
        \caption{{Labelled predicates for the example STL formula.}}
        \label{tab:constraints}
        \begin{tabular}{|l|l|}
            \hline
            {$\|x_1-x_2\|-1$} & {$h^c_{1,1},\ h^c_{2,1}$}\\
            \hline
            {$\|x_2-x_3\|-1$} & {$h^c_{2,2},\ h^c_{3,1}$}\\
            \hline
            {$\|x_3-x_4\|-1$} & {$h^c_{3,2},\ h^c_{4,1}$}\\
            \hline
            {$\|x_4-x_1\|-1$} & {$h^c_{4,2},\ h^c_{1,2}$}\\
            \hline
            {$\|x_1\|-1$} & {$h^d_{1,1}$}\\
            \hline
            {$\|x_2-x_3-x_4\|-1$} & {$h^c_{2,3},\ h^c_{3,3},\ h^c_{4,3}$}\\
            \hline
            {$\|x_2\|-1$} & {$h^d_{2,1}$}\\
            \hline
            {$\|x_5\|-1$} & {$h^d_{5,1}$}\\
            \hline
        \end{tabular}
    \end{table}
The subscript $i$ in the labels of the predicate functions $h^d_{i,l}$ and $h^c_{i,j}$ specifies the robot responsible for the predicate function.
\end{exmp}
Now we are ready to define the communication structure of the multi-robot system, which is dictated by a graph. 
Let the communication graph be given by $\mathbf{G}=(\mathbf{V},\mathbf{E})$ where $\mathbf{V}$ is the set of vertices corresponding to the indices of the robots and $\mathbf{E}$ is the set of edges. In particular, the edge $(i,j)\in\mathbf{E}$ indicates that robot $i$ can communicate with robot $j$ as the subsequent assumption states. 
\begin{assumption}\label{ass:communication}
 If $(i,j)\in\mathbf{E}$, then robots $i$ and $j$ can continuously communicate with each other.
\end{assumption}
We further assume that, every coupled predicate function induces an edge, ensuring that all state variables needed to compute the predicate function are locally available for each robot. 
\begin{assumption}\label{ass:graph_edge}
  If there exists $k\in\{1,\dots,K\}$ such that $i,j \in\mathbf{S}_k$, for $i\neq j$, then $(i,j)\in\mathbf{E}$.
\end{assumption}
Note that the aforementioned assumption implies that the graph is undirected i.e., $(i,j)\in\mathbf{E}$ implies $(j,i)\in\mathbf{E}$. Additionally, based on $\mathbf{G}$, the neighbourhood set $\mathcal{N}_i$ of a robot $i$ is defined as $\mathcal{N}_i=\{j\in\mathbf{V}\ |\ (i,j)\in\mathbf{E}\}$. {We also assume that $\mathbf{G}$ is static, meaning that no new vertices are added and no edges are created or deleted.}
With the above assumptions, we are ready to define the distributed information flow:
\begin{defn}
    {An algorithm is called distributed, if it can be executed individually by each robot $i$ (a local version of the algorithm) by using only information from its neighbours $\mathcal{N}_i$.}
\end{defn}
This definition of distributed algorithm does not allow for any global information sharing among robots that are not neighbours with each other, thus a central computer cannot evaluate an STL formula. For example, consider the STL formula:
\[
\varphi = \Big(\|\mathbf{x}_1-\mathbf{x}_2\|\leq 1\Big)\ \land\ \Big(\|\mathbf{x}_2-\mathbf{x}_3\|\leq 1\Big)
\]
where $\mathbf{x}_1$, $\mathbf{x}_2$, and $\mathbf{x}_3$ are the states of robot $1$, $2$, and $3$ respectively. 
Then, Assumption \ref{ass:graph_edge} allows for a communication link between robot $1$ and robot $2$, and between robot $2$ and robot $3$. A distributed algorithm, in the sense of our work, does not allow for communication between robot $1$ and robot $3$.\\
}
We are now ready to formally state the problem addressed in this paper. 
\begin{prob}\label{problem}
Given an STL formula $\varphi$ that specifies tasks on a multi-robot system with $\mathbf{N}$ robots, design a distributed algorithm to find continuous time-varying trajectories $\mathbf{y}_i:[0,\mathrm{th}(\varphi)]\to \mathcal{W}_i$, starting at an initial configuration $\mathbf{y}_i(0)=\mathbf{x}_i(0)$, such that $(\mathbf{y},t)\models \varphi,\ \forall t\in[0,\mathrm{th}(\varphi)]$ with $\mathbf{y}:=[\mathbf{y}_1^\top,\mathbf{y}_2^\top,\dots,\mathbf{y}_{\mathbf{N}}^\top]^\top$.
\end{prob}
It should be noted that we do not currently address the closed-loop stability of the underlying multi-robot system. Instead, we focus on the trajectory generation aspect and rely on existing low-level control approaches to track the generated trajectories. For more information, see Remark \ref{rmrk:dynamic}. The above problem is addressed assuming that at least one such solution exists. This will help us provide probabilistic completeness guarantees later on. Formally, we state the following assumption:
\begin{assumption}\label{ass:feasible}
    There exists at least one $\mathbf{y}$ such that $(\mathbf{y},t)\models \varphi$.
\end{assumption}

\section{STL Formula Decomposition}\label{sec:decompose}

In this section, we present how to retrieve spatial and temporal constraints from a given STL formula $\varphi$. %We start by presenting the multi-robot system and defining the types of constraints, followed by presenting the topology and finally stating the problem being addressed. 
\subsection{Spatial Constraints}
{In Section \ref{subsec:problem}, we provisioned each predicate function $h^{(k)}(\mathbf{x})$ appearing in the STL formula $\varphi$ over the complete multi-robot system to the corresponding robot $i$, denoting them as $h^d_{i,l}$ and $h^c_{ij}$, depending om whether the robot is responsible for an independent or a coupled task.}

For robot $i$, cast the constraints \eqref{eq:labelled_constraint_predicates} into the cost function $F^i$ as 
\begin{equation}\label{optimisation_func}
    F^i := \sum_{l=1}^{L_i} \frac{1}{2}\max\Big(0,{ h^d_{i,l}}\Big)^2 +  \sum_{\substack{j \in \{1,\dots,K_c\} \\ :i \in \mathbf{S}_j}} \frac{1}{2} \max\Big(0, h^c_{i,j}\Big)^2
\end{equation}
Observe that $F^i:\mathcal{W}\to \mathbb{R}_{+}$ and $F^i=0$ if and only if all the constraints in \eqref{eq:labelled_constraint_predicates} are satisfied. Then, enforcing conditions \eqref{eq:labelled_constraint_predicates} is equivalent to finding $\mathbf{x}_i$ for a given  $\mathbf{x}_j\ (j\in\mathcal{N}_i)$ such that $F^i=0$. This problem can be posed as,
\begin{equation}\label{tro:eq:optimisation}
%\min_{\mathbf{x}_i\in \mathcal{X}_i\cap \mathcal{X}_{ij}} F^i
\min_{\mathbf{x}_i\in \mathcal{W}_i} F^i
\end{equation}
whose solution $\mathbf{x}_i^\star$ satisfies $F^i(\mathbf{x}^\star)=0$. In the cost function \eqref{optimisation_func}, to reduce computational costs, we only minimise $h^{(k)}(\mathbf{x}(t))$ when $h^{(k)}(\mathbf{x}(t))>0$ while leaving $h^{(k)}(\mathbf{x}(t))\leq 0$ unchanged. This leads us to minimise: $F^i = \max(0,h^{(k)}(\mathbf{x}(t)))$, which results in $F^i=h^{(k)}(\mathbf{x}(t))$ when $h^{(k)}(\mathbf{x}(t))>0$ and $F^i=0$ when $h^{(k)}(\mathbf{x}(t))\leq 0$. Additionally, squaring the function penalises larger errors more than smaller ones. Other cost functions that enforce $h^{(k)}(\mathbf{x}(t))\leq 0$ within the validity domain can also be considered. For example, the cost function  $F^i := \sum_{l=1}^{L_i} { h^d_{i,l}}^2 +  \sum_{j} { h^c_{i,j}}^2$, $j\in\{1,\dots,K_c\}:i\in\mathbf{S}_j$ could also be used. However, it was not our first choice, as we aimed to minimise $h^{(k)}(\mathbf{x}(t))$ only when $h^{(k)}(\mathbf{x}(t))>0$, whereas this cost function would attempt to minimise $h^{(k)}(\mathbf{x}(t))$ regardless of its sign. Additionally, our formulation is general and works for any type of STL formula as any type of objectives can be encoded in the STL formula, from which, we can extract the predicate functions and minimise the function  $F^i$.

{
The solution for finding the global minimum of a nonconvex function is a subject of extensive research. We argue that employing gradient descent with random initialisations is adequate for addressing this problem, particularly since the initialisations are sampled from a compact set, $\mathcal{W}_i$. Furthermore, using the knowledge that the minimum of the function, $F^i(\mathbf{x})=0$, acts as a stopping criterion and facilitates the attainment of the desired solution. We direct readers to the seminal work in \cite{4749425}, which presents a distributed gradient descent algorithm for multi-agent systems. Additionally, under certain assumptions, \cite{daneshmand} demonstrates that gradient descent with a constant step size avoids entrapment at saddle points. Gradient descent is also shown to efficiently manage most reach-avoid constraints without the need for re-initialisation, given that such constraints are expressible using norms. }For our application of gradient descent, we utilise the function presented in Function \ref{tro:alg:optimisation}.
\renewcommand{\algorithmcfname}{Function}
\begin{algorithm}
    %\noindent\textbf{Function 1:} Special Optimization Function
    \caption{DistributedOptimisation}    
    \label{tro:alg:optimisation}
    \SetKwFunction{Optimisation}{Optimisation}
    \SetKwFunction{GC}{GradientComputation}
    \KwIn{$\mathbf{x}_i, \text{step size}\ \delta,$ $\text{maximum iterations}\ L^\prime,$ $\text{activation variables } \lambda_{ij}, k\gets0$ }
    \KwOut{$\mathbf{x}_i^\star$}
    Receive neighbour states $\mathbf{x}_m$ for all $m\in\mathcal{N}_i$\;
    Compute $F^i$ with weights $\lambda_{ij}$\;
    %$F^i=\sum_j \lambda_{ij}\max(0,h_{ij})^2$\;
    $\nabla F^i \gets$ \GC{$\mathbf{x}_i^{\text{inter}}, \mathbf{x}_j^{\text{neigh}}$}\;
    $\Delta \mathbf{x}_i = - \nabla F^i$\;
    \While{$ F^i> 0$\label{tro:line:while_gradient1}}
    {
    $\mathbf{x}_i := \mathbf{x}_i + \delta \Delta \mathbf{x}_i$\;
    Receive neighbour states $\mathbf{x}_m$ for all $m\in\mathcal{N}_i$\;\label{tro:line:communication21}
    $\nabla F^i \gets$ \GC{$\mathbf{x}_i, \mathbf{x}_j$}\;\label{line_gc}
    $\Delta \mathbf{x}_i = - \nabla F^i$\;\label{tro:line:while_ends_gradient1}
    $k\gets k+1$\;
    \lIf{$k>L^\prime$}{\texttt{random} $\mathbf{x}_i\in\mathcal{W}_i$ \Break}\label{tro:line:gradient_break1}
    }
\end{algorithm}
\renewcommand{\algorithmcfname}{Algorithm}
Function \ref{tro:alg:optimisation} implements the gradient descent algorithm as described in Algorithm 9.3 of \cite{boyd_vandenberghe_2004}, utilising initial conditions $\mathbf{x}_i$, step size $\delta$, maximum number of iterations $L'$, and activation variables  $\lambda_{ij}$ as inputs. The activation variables are presented later in Function \ref{tro:alg:GradientDescent}. It returns the optimised states $\mathbf{x}_i^\star$ as output. In line \ref{line_gc}, the function \texttt{GradientComputation()} computes the gradient, either analytically or numerically. {The stopping criterion is met either when a feasible state is determined, indicated by $F^i= 0$, or when the iteration count exceeds $L'$ (line \ref{tro:line:gradient_break1}), which may occur due to multiple conflicting predicates active within $F^i$. {This situation arises because the algorithm accounts for the possibility that the \textit{eventually} operator may not be satisfied at every sampled point within its validity domain. This occurs, for example, if $\varphi=\mathcal{F}_{[0,5]}\mathcal{G}_{[0,5]}\ (g^{(1)}(\mathbf{x})\leq \epsilon_1)\ \land \mathcal{G}_{[5,10]}\ (g^{(2)}(\mathbf{x})\leq \epsilon_2)$, and there is a conflict between $h^{(1)}(\mathbf{x}):=g^{(1)}(\mathbf{x})- \epsilon_1$ and $h^{(2)}(\mathbf{x}):=g^{(2)}(\mathbf{x})- \epsilon_2$ (i.e., $\nexists \mathbf{x}_i^\star\in\mathcal{W}_i$ such that $h^{(1)}(\mathbf{x})(\mathbf{x}_i^\star)\leq 0 \land h^{(2)}(\mathbf{x})(\mathbf{x}_i^\star)\leq 0$). In such cases, it becomes necessary for $h^{(1)}(\mathbf{x})\leq 0$ to be true exclusively within the interval $[0,5]$[s] and for $h^{(2)}(\mathbf{x})\leq 0$ to be true exclusively within the interval $[5,10]$[s].}}
%In lines \ref{tro:line:while_gradient}-\ref{tro:line:while_ends_gradient}, we implement the standard gradient descent algorithm with a step 

The robots solve their respective optimisation problem cooperatively in a distributed manner via inter-neighbour communication. This makes the problem distributed, as every interaction between robots is part of the communication graph. Given the nature of the optimisation problem, there is a trade-off between robustness and optimisation performance since $\mathbf{x}^\star$ converges to the boundaries imposed by the STL formula constraints, making it vulnerable to potential perturbations. However, introducing a slack variable into the equation can enhance robustness, albeit at the cost of sacrificing completeness arguments. The example below shows how to construct the optimisation functions $F^i$.
\begin{exmp}Consider a system with 3 agents and the corresponding states $\{x_1,x_2,x_3\}$, and let the STL formula be: $\varphi = (\|x_1-x_2\|>5)\ \land (\|x_2-x_3\|<2)$; then, the functions $F^i$, for $i\in\{1,2,3\}$, are,
\begin{align*}
F^1 &= \frac{1}{2}\max(0,5-\|x_1-x_2\|)^2\\
F^2 &= \frac{1}{2}\max(0,5-\|x_1-x_2\|)^2 + \frac{1}{2}\max(0,\|x_2-x_3\|-2)^2\\
F^3 &= \frac{1}{2}\max(0,\|x_2-x_3\|-2)^2.
\end{align*}
\end{exmp}

Now that spatial constraints are encoded into the optimisation problem, we are ready to encode temporal constraints in the following section, thus completing our STL decomposition into spatial and temporal constraints. %Note that we use $\mathbf{x}^i$ and $\mathbf{x}_i$ interchangeably, as convenient in the context, to represent state of robot $i$. 
\subsection{Temporal Constraints}\label{subsec:validity_domain}

We now introduce the concept of \emph{validity domain}, a time interval associated with every predicate and defined for every path in the STL formula. This interval represents the time domain over which each predicate applies and is defined as follows:
{\begin{defn}\label{defn:validity_domain}
The \emph{validity domain} $\mathrm{vd}(\bar{\varphi})$ of each path $\bar{\varphi}$ of an STL formula $\varphi$, is recursively defined as
\begin{equation}
    \mathrm{vd}(\bar{\varphi}) = 
    \begin{cases} 
        0,& \text{if}\ \bar{\varphi}=\mu^h\\ 
    \mathrm{vd}(\bar{\varphi}_1),& \text{if}\  \bar{\varphi}=\neg\bar{\varphi_1}\\ 
    [a,b] ,& \text{if}\  \bar{\varphi} = \mathcal{G}_{[a,b]}\mu^h\\ 
    a \oplus\mathrm{vd}(\bar{\varphi}_1), & \text{if}\  \bar{\varphi}=\mathcal{G}_{[a,b]}\bar{\varphi}_1,\bar{\varphi}_1\neq\mu^h\\
    t^\star + T^\star\oplus\mathrm{vd}(\bar{\varphi}_1), & \text{if}\  \bar{\varphi}=\mathcal{F}_{[a,b]}\bar{\varphi}_1
    \end{cases}
\end{equation}
\noindent where $T^\star:=\{t\in[a,b]\ |\ (\mathbf{x},t)\models \mathcal{F}_{[a,b]}\bar{\varphi}\}$ is a time instant in $[a,b]$ when the state $\mathbf{x}$ evaluated at $t$ of a signal $\mathbf{x}(t)$ satisfies the eventually operator. The variable %$t^\star = \{t\ |\ (\bar{\varphi},t)\models \mathcal{F}_{[a,b]}\bar{\varphi}_1\}$ 
$t^{\star}$ is initialised to 0, but takes the value $t^\star = T^\star$ every time $T^{\star}$ is updated and thus captures the last instance of satisfaction for the eventually operator. 
\end{defn}
}
The above definition of $t^\star$ is necessary due to the redundancy of the eventually operator; we must ascertain the specific instances where the eventually condition is met to ensure finding a feasible trajectory. 
Additionally, we need to maintain the history of $T^\star$ for nested temporal operators which require recursive satisfaction.
 The validity domain is determined for each path of an STL formula in a hierarchical manner, beginning at the root of the tree, and each path has a distinct validity domain. The number of leaf nodes in an STL formula is equal to the total number of validity domains. In Definition \ref{defn:validity_domain}, we do not include the operators $\land$ and $\lor$ because they do not impose temporal constraints on the predicates and thus inherit the validity domains of their parent node. If there is no parent node, operators $\land$ and $\lor$ inherit the validity domains of their child node.

\begin{rmrk}\label{remark:pathological}
The validity domain is specially defined in the following cases. If a path contains only predicates, the validity domain of $\mu^h$ is equal to the time horizon of $\varphi$ (i.e., $\mathrm{vd}(\mu^h)=\mathrm{th}(\varphi)$). Furthermore, if a path contains nested formulas with the same operators, such as $\bar{\varphi}=\mathcal{G}_{[1,10]}\mathcal{G}_{[0,2]}(\cdot)$, then the validity domain of $\bar{\varphi}$ is equal to the time horizon of the path (i.e., $\mathrm{vd}(\bar{\varphi})=\mathrm{th}(\bar{\varphi})$). For example, $\mathrm{vd}({\mathcal{G}_{[1,10]}\mathcal{G}_{[0,2]}(\cdot)})=\mathrm{th}(\bar{\varphi})=[1,12]$.
\end{rmrk}

\begin{exmp}
Consider the following examples of the validity domain:
\begin{itemize}
    \item $\varphi_1 = \mathcal{G}_{[5,10]}(g^{(1)}(\mathbf{x})\leq \epsilon_1)$, then $\mathrm{vd}(\varphi_1)=[5,10]$, which is the interval over which $\mu^{h^{(1)}}$ must hold. Here $\mu^{h^{(1)}}$ is the predicate corresponding to the predicate function $h^{(1)}(\mathbf{x})=g^{(1)}(\mathbf{x})-\epsilon_1$.
    \item $\varphi_2 = \mathcal{F}_{[5,10]}(g^{(1)}(\mathbf{x})\leq \epsilon_1)$, then $t^{\star}$ is initialised to $0$, $T^\star\in[5,10]$ and $\mathrm{vd}(\mu^{h^{(1)}})=0$. Therefore, $\mathrm{vd}(\varphi_2)=T^{\star}\in [5,10]$ is the instance when $\mu^{h^{(1)}}$ must hold. 
    \item $\varphi_3 = \mathcal{F}_{[5,10]}\mathcal{G}_{[0,2]}(g^{(1)}(\mathbf{x})\leq \epsilon_1)$, then $t^\star$ is initialised to $0$, $T^\star \in[5,10]$, $\mathrm{vd}(\mathcal{G}_{[0,2]}(g^{(1)}(\mathbf{x})\leq \epsilon_1)) = [0,2]$. Therefore, $\mathrm{vd}(\varphi_3)=0+T^\star\oplus[0,2]=[T^\star,T^\star+2]$ is the interval over which $\mu^{h^{(1)}}$ must hold such that $\varphi_3$ is satisfied. 
    \item $\varphi_4 = \mathcal{G}_{[2,10]}\mathcal{F}_{[0,5]}(g^{(1)}(\mathbf{x})\leq \epsilon_1)$, then $a = 2$ and $\mathrm{vd}(\varphi_4) = 2 \oplus \mathrm{vd}(\mathcal{F}_{[0,5]}(g^{(1)}(\mathbf{x})\leq \epsilon_1)) = 2 + 0 + T^\star$ where $T^\star\in[0,5]$. Suppose $T^\star = 1$, then $\mathrm{vd}(\varphi_4)=3$ is the time instance when $\mu^{h^{(1)}}$ must hold. Once $\mu^{h^{(1)}}=\top$, then $t^\star = T^\star$ and the new $\mathrm{vd}(\varphi_4)=2 + 1 + T^\star$ where $T^\star\in [0,5]$. 
    \item $\varphi_5 = \mathcal{F}_{[0,100]}\mathcal{G}_{[5,10]}\mathcal{F}_{[0,1]}(g^{(1)}(\mathbf{x})\leq \epsilon_1)$, then $t^\star =0$, $T^\star \in[0,100]$ and $\mathrm{vd}(\varphi_5)=T^\star + a \oplus \mathrm{vd}(\mathcal{F}_{[0,1]}(g^{(1)}(\mathbf{x})\leq \epsilon_1))$. Suppose $T^\star = 50$, then $\mathrm{vd}(\varphi_5)= 55 \oplus \mathrm{vd}(\mathcal{F}_{[0,1]}(g^{(1)}(\mathbf{x})\leq \epsilon_1))$ and so on.
\end{itemize}
Regarding the STL formula in equation \eqref{eq:stl_eg}, the validity domains are defined for the following paths: $ \mathcal{F}_{I_1}\rightarrow \mu^{h^1},\  \mathcal{F}_{I_1}\rightarrow \mathcal{G}_{I_2}\rightarrow \mu^{h^2},\ \mathcal{G}_{I_3}\rightarrow\mathcal{F}_{I_4}\rightarrow\mu^{h^3}$, and $\mathcal{G}_{I_5}\rightarrow\mu^{h^4}$. 
\end{exmp}

We use the following notational convenience in this work: if a parent node of a leaf node of a path $\bar{\varphi}$ is an \emph{eventually} operator we denote the corresponding validity domain by $\mathrm{vd}^F()$, and, if the parent node of a leaf node of a path $\bar{\varphi}$ is an \emph{always} operator we denote the corresponding validity domain by $\mathrm{vd}^G()$.  The notation $\mathrm{vd}^F()$ indicates that the predicate of the respective leaf node needs to hold at some instance in the said interval, and $\mathrm{vd}^G()$ indicates that the predicate of the respective leaf node needs to hold throughout the interval. The following lemma formalises the relation between the STL formula and its corresponding encoding as described above. 

{
\begin{lemma}\label{lemma:constraints}
%Let $\mathbf{x}(t)=[\mathbf{x}_1^\top, \mathbf{x}_1^\top,\dots]$ denote the trajectories of all robots and let $\bar{\varphi}_k$  be all the subformulas corresponding to the STL formula $\varphi$. Then, if $\sum F^i(\mathbf{x}(t))\leq 0$ for all $t\in \bigcup_k \mathrm{vd}(\bar{\varphi}_k)$ implies $\mathbf{x}(t)\models\varphi$.
Suppose $\mathbf{x}(t)=[\mathbf{x}_1^\top, \mathbf{x}_2^\top,\dots]$ represents the states of all robots, and $\bar{\varphi}_k$ encompasses all subformulas associated with the STL formula $\varphi$. %If for all $t$ in $\bigcup_k \mathrm{vd}(\bar{\varphi}_k)$, if $\sum_i F^i(\mathbf{x}(t))\leq 0$ holds, then it follows that $\mathbf{x}(t)\models\varphi$.
Let $\sum_i F^i(\mathbf{x}(t))= 0$ for all $t\in \bigcup_k \mathrm{vd}(\bar{\varphi}_k)$. Then, it holds that $\mathbf{y}(t)\models\varphi$.
%Given a trajectory $\mathbf{x}(t)$ and an STL formula $\varphi$. If 
%$\mathbf{x}(t)$ satisfies the constraints encoded in Sections \ref{subsec:distributed_optimisation} and \ref{subsec:validity_domain}, then $\mathbf{x}(t)\models \varphi$.
\end{lemma}
\begin{proof}
The proof follows from the construction of the optimisation function \eqref{optimisation_func} and the validity domain. Notice that if the optimisation problem \eqref{tro:eq:optimisation} converges to the desired minima at $F^i(\mathbf{x})=0$, then $\mu^{ h^d_{i,l}} = \top$ and $\mu^{ h^c_{i,j}} = \top$ for all $l\in\{1,\dots,L_i\}$ and $j\in\{1,\dots,K_c\}:i\in\mathbf{S}_j$. Next, by definition, the validity domain is defined for the STL formula and if $F^i$ is minimised during the validity domain, then $\mathbf{y}(t)\models\varphi$. \\
\end{proof}
}

In the next Section, we present how to integrate the validity domain with the optimisation problem in \eqref{tro:eq:optimisation}, completing thus the spatial and temporal integration.  

%%%%%%%%%%%%%%%%%%%%%%%%%%%%%%%%%%%%%%%%%%%%%%%%%%%%%%%%%%%%%%%%%%%%%%%%%%%%%%%%
%%%%%%%%%%%%%%%%%%%%%%%%%%%%%%%%%%%%%%%%%%%%%%%%%%%%%%%%%%%%%%%%%%%%%%%%%%%%%%%%

\section{Main Results}\label{sec:main}

\begin{figure*}
     %\centering
     \begin{subfigure}{0.53\columnwidth}
         %\centering
         \includegraphics[width=\columnwidth]{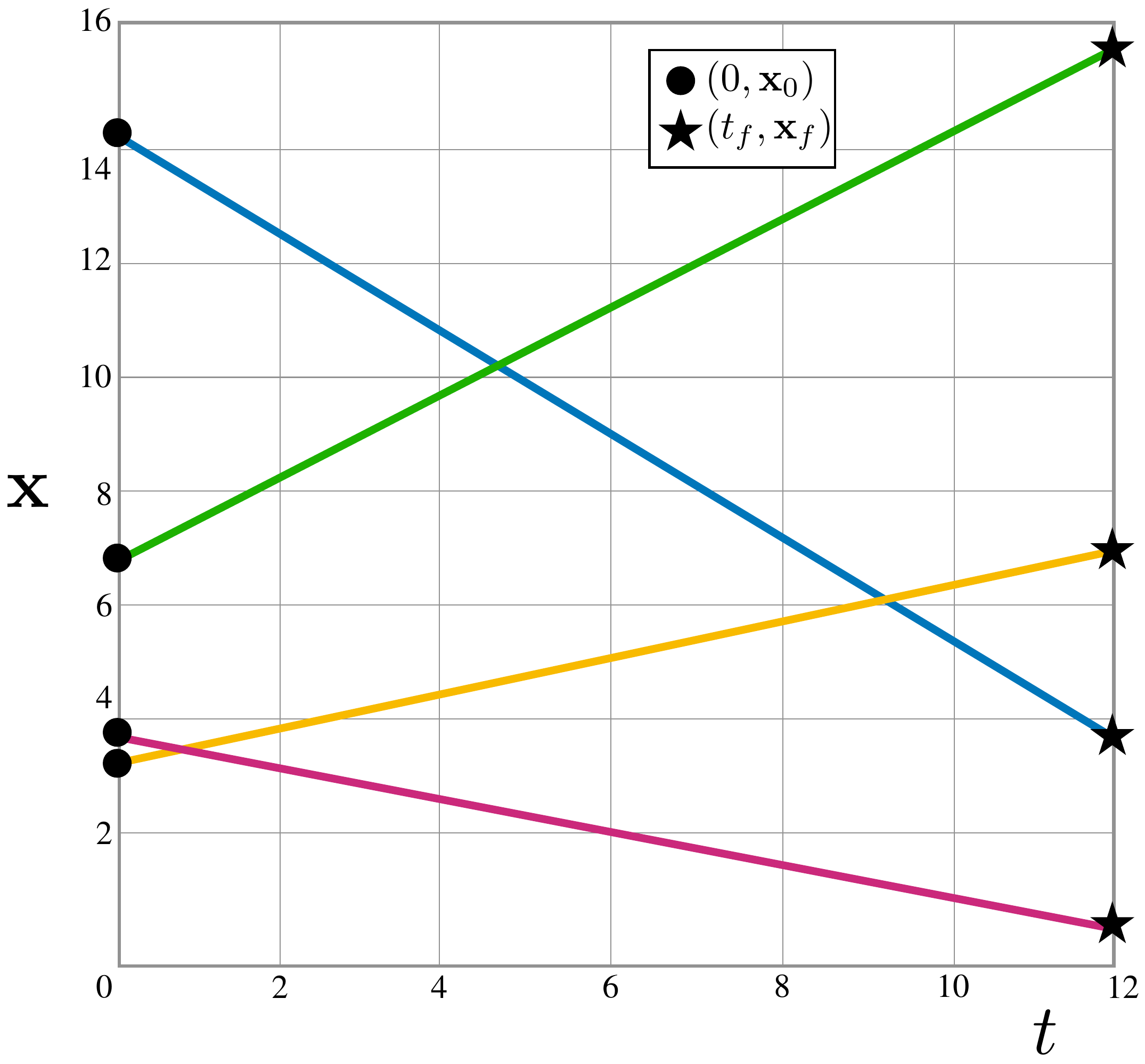}
         \subcaption{Initial trajectory.}
         \label{tro:fig:step1}
     \end{subfigure}%
     \begin{subfigure}{0.5\columnwidth}
         %\centering
         \includegraphics[width=\columnwidth]{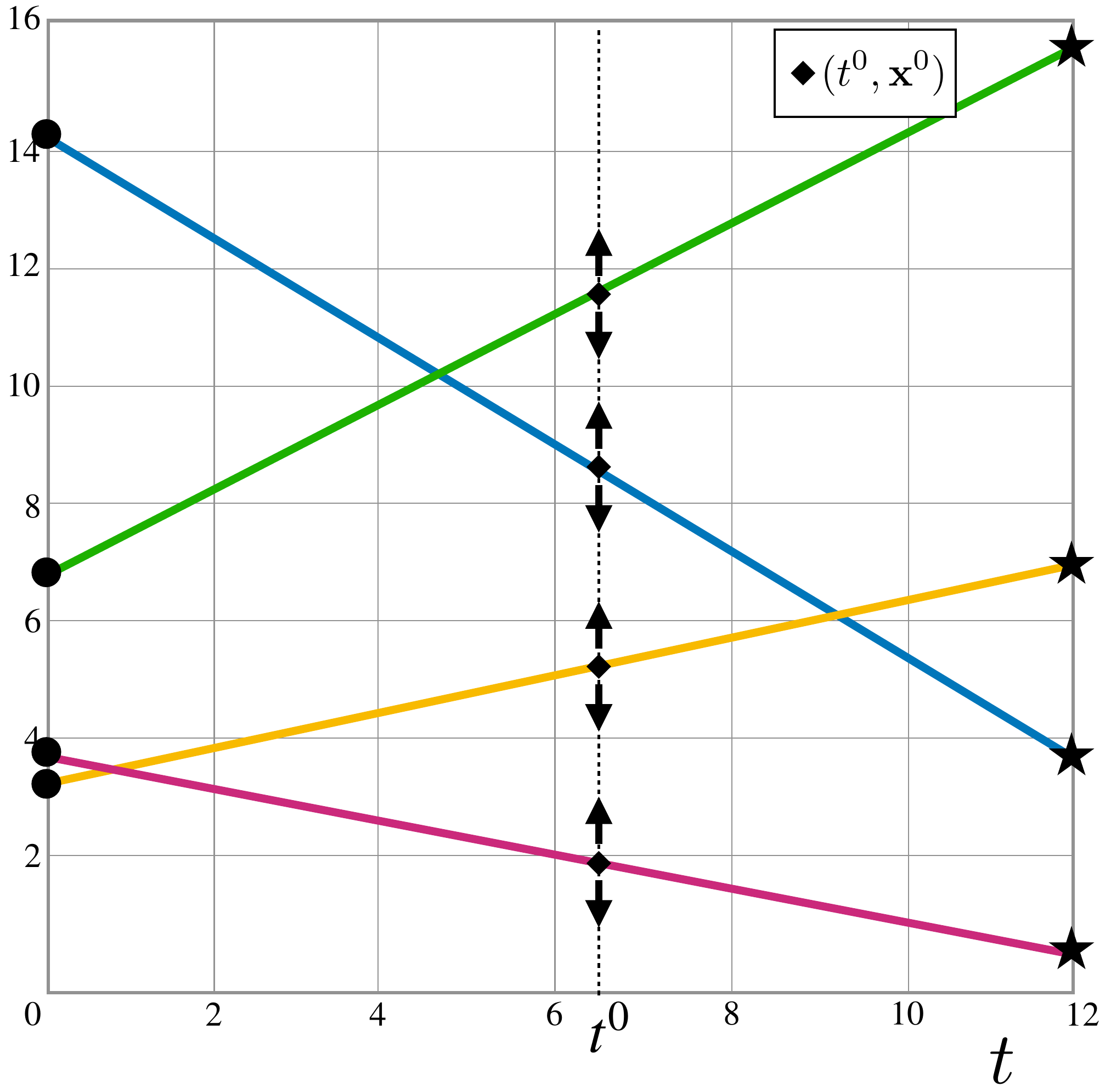}
         \subcaption{A random time instance $t^0$. }
         \label{tro:fig:step2}
     \end{subfigure}
     \begin{subfigure}{0.5\columnwidth}
         %\centering
         \includegraphics[width=\columnwidth]{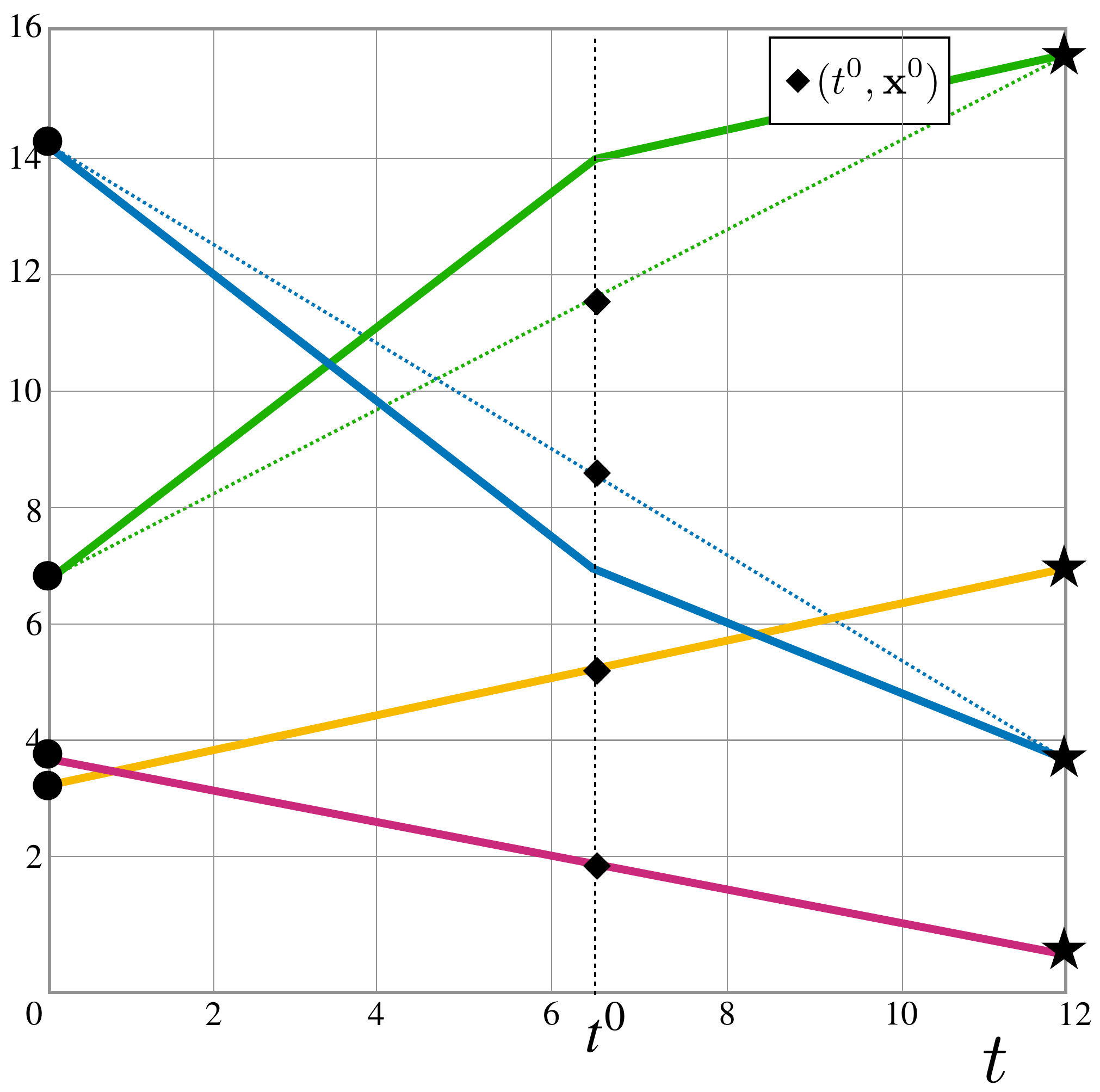}
         \subcaption{Optimal solution $\mathbf{x}^{\star}$.}
         \label{tro:fig:step3}
     \end{subfigure}%
     \begin{subfigure}{0.5\columnwidth}
         %\centering
         \includegraphics[width=\columnwidth]{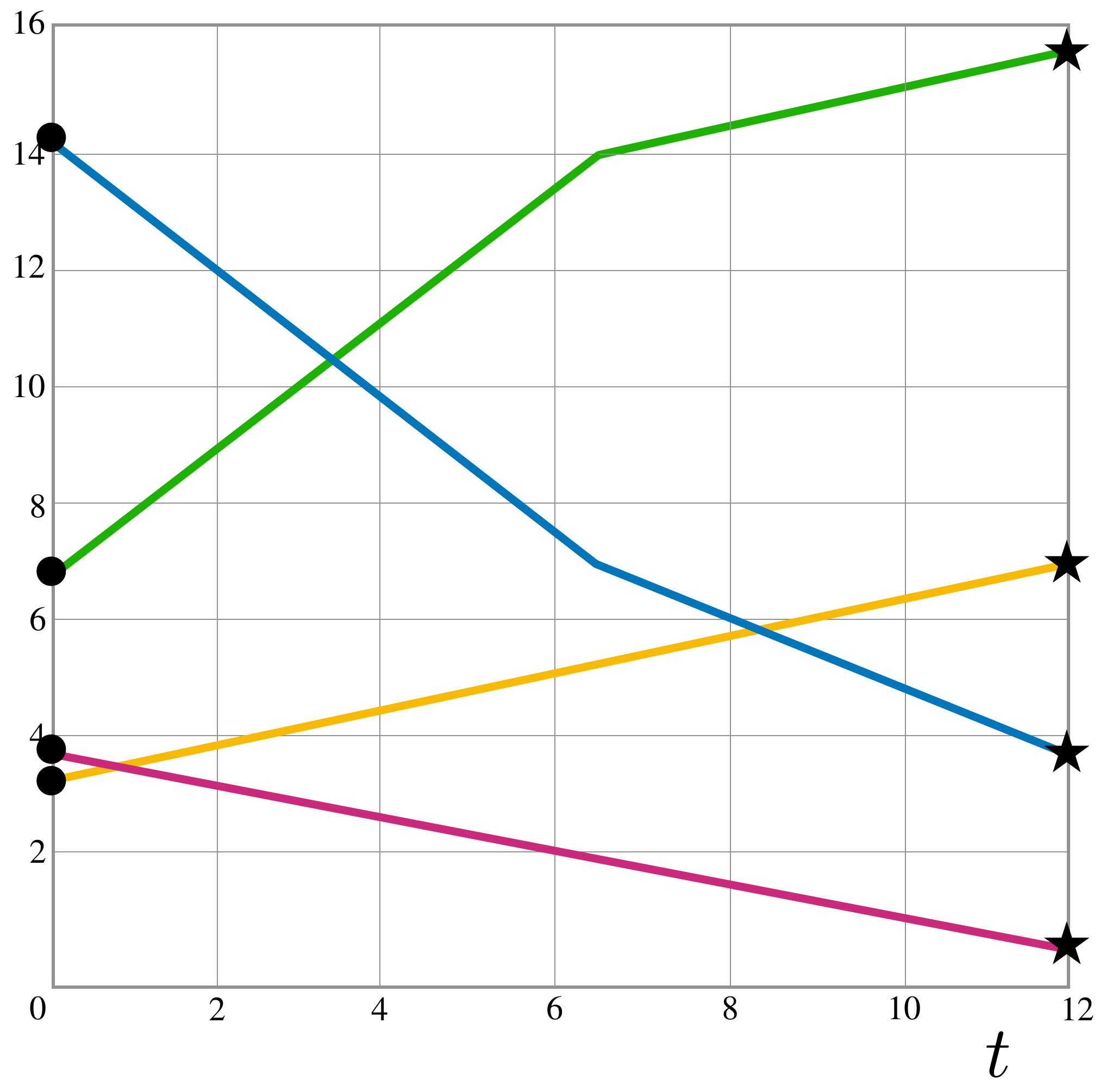}
         \subcaption{Shaped trajectory.}
         \label{tro:fig:step4}
     \end{subfigure}%
        \caption{Illustration of the proposed algorithm.}
        \label{fig:illustration}
\end{figure*}

In this section, we present the algorithm for generating continuous trajectories that meet the requirements of a given STL formula $\varphi$. {The algorithm is executed by the robots offline in a distributed manner, in the sense that they only communicate with their neighbouring robots.} The algorithm builds a tree $\mathcal{T}_i=\{\mathcal{V}_i, \mathcal{E}_i\}$ for robot $i$ where $\mathcal{V}_i$ is the vertex set and $\mathcal{E}_i$ is the edge set. Each vertex $z\in \mathbb{R}_+\times \mathcal{W}_i$ is sampled from a space-time plane. Until now, we denoted the states of robot $i$ as $\mathbf{x}_i$, but from here onward, we denote them as $\mathbf{x}^i$. 

%%%%%%%%%%%%%%%%%%%%%%%%%%%%%%%%%%%%%%%%%%%%%%%%%%%%%%%%%%%%%%%%%%%%%%%%%%%%%%%%

In what follows, we give a high-level description of the algorithm. The general idea is to start with an initial trajectory that spans the time horizon of the formula $\mathrm{th}(\varphi)$, then repeatedly sample random points along the trajectory and use gradient-based techniques to find solutions that satisfy the specification at these points. More specifically, the algorithm begins by connecting the initial and final points $z_0^i=\{t_0^i,\mathbf{x}_0^i\}$ and $z_f^i=\{t_f^i,\mathbf{x}_f^i\}$ with a single edge $\mathcal{E}_i=\{(z_0^i,z_f^i)\}$.  The initial conditions $z_0^i=\{t_0^i,\mathbf{x}_0^i\}$ depend on the robot's initial position and time. The final conditions are chosen to be $z_f^i=\{\mathrm{th}(\varphi)+\epsilon, \mathbf{x}_f^i\}$ where $\epsilon>0$ and $\mathbf{x}_f^i\in\mathcal{W}^i$. Let $t_0^i=0$ and $t_f^i=\mathrm{th}(\varphi)+\epsilon$. The final states $\mathbf{x}_f^i$ can be randomly chosen since the states in the interval $[0,\mathrm{th}(\varphi)]$ will be determined by the algorithm based on the constraints imposed by $\varphi$. The algorithm then randomly selects a time instant $t^0\in[0,\mathrm{th}(\varphi)]$ and uses linear interpolation to determine the states of each robot at that time, denoted by $\mathbf{x}^0$. The robots then solve the distributed optimisation problem \eqref{tro:eq:optimisation} to find new positions $\mathbf{x}^\star$ that meet the specification at time $t^0$. The algorithm then repeats this process at a user-specified time density, updating the trajectories as necessary. The result is a trajectory that asymptotically improves the task satisfaction of the STL formula. 

%The result is a continuously changing trajectory that satisfies the STL formula $\varphi$.
\begin{exmp}

Before we get into the technical details, let us consider an example of 4 agents, represented by the colours \emph{blue, green, yellow} and \emph{magenta}, to illustrate the procedure. 
Suppose, at a specific instance in time, say $t^0$, the STL formula requires agent 1 (\textcolor{blue}{blue}) and agent 2 (\textcolor{green}{green}) to be more than 6 units apart and agent 3 (\textcolor{yellow}{yellow}) and agent 4 (\textcolor{magenta}{magenta}) to be closer than 6 units i.e., for $\epsilon>0$, \begin{align*}G_{[t^0-\epsilon, t^0+\epsilon]}\Big((\text{\textcolor{blue}{blue} and \textcolor{green}{green} are farther than 6 units apart})\land \\ \text{(\textcolor{yellow}{yellow} and \textcolor{magenta}{magenta} are closer than 6 units)}\Big)\end{align*} 
We begin the process by connecting the initial and final points $z_0^i$ and $z_f^i$ with an initial trajectory for all agents, as shown in Figure \ref{tro:fig:step1}. Each agent's vertex set is $\mathcal{V}_i$ and consists of the start and end points denoted by $z_0^i$ and $z_f^i$ respectively, while its edge set $\mathcal{E}_i$ contains only one edge connecting the start and end points. %The initial trajectory of agent $i$ is the tree $\mathcal{T}_i=\{\mathcal{V}_i, \mathcal{E}_i\}$.  
From the initial trajectory, the algorithm randomly selects a point at time instance $t^0$ from the entire time domain and uses linear interpolation to determine the state of each agent at that time. The agents solve \eqref{tro:eq:optimisation} using the initial position $\mathbf{x}^0$ to find new position $\mathbf{x}^\star$, as seen in Figure \ref{tro:fig:step2}. 
As shown in Figure \ref{tro:fig:step3}, the distributed optimisation problem \eqref{tro:eq:optimisation} is solved, resulting in a solution $\mathbf{x}^\star$, in which agent 1 and agent 2 are positioned so that they are more than 6 units apart and agent 3 and agent 4 remain undisturbed. The latter is the result of using functions of the form $1/2 \max(0,h^c_{i,j})^2$, and since agent 3 and agent 4 already satisfy the requirements, i.e., $h^c_{i,j}<0$, the function is valued 0. The newly determined positions of agents 1 and 2 are added to the tree, allowing the trajectory to be shaped to meet the requirements. The updated trajectory can be seen in Figure \ref{tro:fig:step4}. This process of randomly selecting a point in time, determining the state of the agents and updating their positions is repeated for a user-defined number of times $L$, to ensure that the trajectory satisfies the STL formula $\varphi$ throughout the time horizon.
    
\end{exmp}

\begin{figure}
     \centering
     \includegraphics[width=\columnwidth]{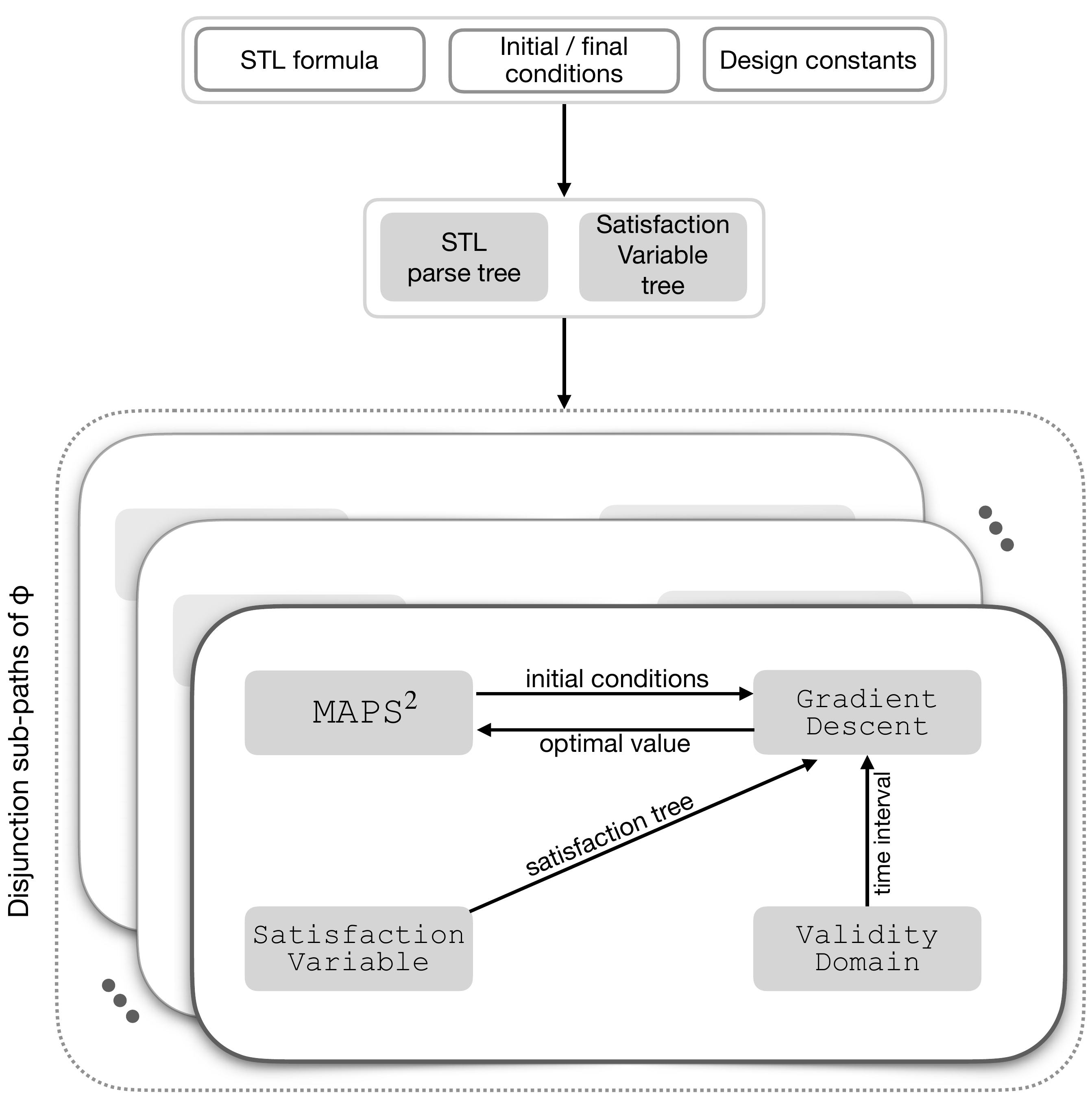}
     \caption{Architecture of the provided algorithm.}
     \label{fig:flow}
\end{figure}%

\subsection{MAPS$^2$}\label{tro:subsec:maps2}
The architecture of the algorithm \texttt{MAPS}$^2$ (short for `\textbf{m}ulti-{r}obot \textbf{a}nytime motion \textbf{p}lanning under \textbf{s}ignal temporal logic \textbf{s}pecifications'), is depicted in Figure \ref{fig:flow}. 
The algorithm, outlined in Algorithm \ref{tro:alg:maps}, begins with an initial trajectory connecting $z_0^i$ and $z_0^f$, along with a random seed and design constants as input (see lines \ref{tro:line:v0}-\ref{tro:line:initial_trajectory}). The random seed ensures that all robots select the same time instance. The algorithm proceeds by repeatedly sampling a time instance within the interval, interpolating states at the said time instance, applying gradient descent to minimise the function \eqref{optimisation_func}, and either adding or discarding the resulting optimal solution. This process is repeated until the total number of vertices, $L$, is reached, see lines \ref{tro:main_for}-\ref{tro:last_line}. This is also illustrated in Figure \ref{fig:illustration}.
\begin{algorithm}
    \SetKwFunction{maps}{MAPS}
    \SetKwFunction{Interpolate}{Interpolate}
    \SetKwFunction{SearchSort}{SearchSort}
    \SetKwFunction{GradientDescent}{GradientDescent}
    \SetKwFunction{ValidityDomain}{ValidityDomain}
    \caption{MAPS$^2$}
    \label{tro:alg:maps}
    \KwIn{Initial condition $z^i_{\text{0}}=\{t^i_{\text{0}}, \mathbf{x}^i_{\text{0}}\}$, Final condition $z^i_\text{f} = \{t^i_{\text{f}}, \mathbf{x}^i_{\text{f}}\}$, Maximum number of nodes $L$, random seed, step size $\delta$, stopping criterion $\eta$, satisfaction variables of all subpaths $\bar{\varphi}_k$: $\tau(\bar{\varphi}_k)\gets-1$}
    \KwOut{$\mathcal{T}_i$}
    $\mathcal{V}_i \gets \mathcal{V}_i \cup z_{\text{0}}^i \cup z_{\text{f}}^i$ \;\label{tro:line:v0}
    $\mathcal{E}_i\gets \mathcal{E}_i \cup \{z_{\text{0}}^i, z_{\text{f}}^i\}$\;\label{tro:line:e0}
    $\mathcal{T}_i\gets \{\mathcal{V}_i, \mathcal{E}_i\}$\;\label{tro:line:initial_trajectory}
    $j\gets 0$\;
    %Set random seed\;
    %$t^0 \gets \text{generate random number in}\ [t^i_{\text{0}},t^i_{\text{f}}]$\;
    %$I\gets$ \VD$(\varphi,\emptyset, t^0)$\;
    \While{$j\leq L$ and $\tau(\textbf{root})\neq +1$}
    %\While{$\tau(\mathrm{\mathbf{root}})\neq +1$}
    {\label{tro:main_for}
    $t^0 \gets \text{generate random number in}\ [t^i_{\text{0}},t^i_{\text{f}}]$\;
    $\text{index} \gets$ \SearchSort{$\mathcal{V}_i,\ t^0$} \;\label{tro:searchsort}
    $z^i_{\text{inter}}\gets$ \Interpolate{$\mathcal{V}_i, index$}\;\label{tro:interpolate}
    $z^i_{\text{opt}}, \tau \gets$ \GradientDescent{$z^i_{\text{inter}},\ \delta,\ L',\ \eta$}\;
    %$ I \gets $\SatisfactionVariable($\pi, t^0, t^0$)\;
    %\lIf{$I>\mathrm{th}(\mu)$}{$I\gets \emptyset$}
    $\mathcal{V}_i \gets \mathcal{V}_i \cup z_{\text{opt}}^i $ \;\label{tro:zopt}
    $\mathcal{E}_i \gets \mathcal{E}_i\ \backslash\ \{z_{\text{index}}^i,\; z_{\text{index+1}}^i\}$\;\label{tro:eremove}
    $\mathcal{E}_i \gets \mathcal{E}_i \cup \{z_{\text{index}}^i,\; z_{\text{opt}}^i\}$\;
    $\mathcal{E}_i \gets \mathcal{E}_i \cup \{z_{\text{opt}}^i,\; z_{\text{index+1}}^i\}$\;\label{tro:eadd}
    $\mathcal{T}_i\gets \{\mathcal{V}_i, \mathcal{E}_i\}$\;\label{tro:last_line}
    $j\gets j+1$\;
    \lIf{$j=L$}{$j\gets0,\ \text{and}\ \forall \mathcal{F}, \tau(\mathcal{F})=-1$}\label{line:maps:tauF}
    }
\end{algorithm}

These steps are implemented as follows: In line \ref{tro:searchsort}, the \texttt{SearchSort()} function separates the vertices $\mathcal{V}_i$ into two sets based on their time values: one set with time values lower than $t^0$ (the vertex with the highest time in this set is indexed with `$\text{index}$'), and another with values greater than $t^0$ (the vertex with the lowest time in this set is indexed with `$\text{index}+1$'). The corresponding vertices are $z^i_{\text{index}} = \{t^i_{\text{index}}, \mathbf{x}^i_{\text{index}}\}$ and $z^i_{\text{index}+1} = \{t^i_{\text{index}+1}, \mathbf{x}^i_{\text{index}+1}\}$.
%The time complexity of this function is $\mathcal{O}(n)$, where $n$ represents the length of the input array.  
Then, the algorithm uses linear interpolation in line \ref{tro:interpolate} via the function \texttt{Interpolate()} to obtain the vertex $z^i_{\text{inter}} = \{t^0, \mathbf{x}^i_{\text{inter}}\}$. This is obtained by solving for $\mathbf{x}^i_{\text{inter}}$ element-wise as the solution of
\[
\mathbf{x}^i_{\text{inter}} =\Big(\frac{\mathbf{x}^i_{\text{index+1}}-\mathbf{x}^i_{\text{index}}}{t^i_{\text{index+1}}-t^i_{\text{index}}}\Big)(t^0-t^i_{\text{index}})+\mathbf{x}^i_{\text{index}}.
\]
The vertex $z_\text{inter}^i$ is the initial condition to solve the optimisation problem \eqref{tro:eq:optimisation}; and once a solution $z_{\text{opt}}^i$ is obtained, it is added to the vertex set $\mathcal{V}_i$ in line \ref{tro:zopt}. The edge set $\mathcal{E}_i$ is reorganised to include $z_{\text{opt}}^i$ in lines \ref{tro:eremove}-\ref{tro:eadd}. 

Moreover, as a safeguard, if a solution remains undiscovered following $L$ iterations, line \ref{line:maps:tauF} initiates a reset procedure. This involves setting the satisfaction variable for all \textit{eventually} operators back to $-1$ and restarting the search.
Since we assume that at least one viable solution always exists (refer to Assumption \ref{ass:feasible}), the absence of a solution occurs solely when an \textit{eventually} operator is satisfied at an impractical instance of time. Such an impractical instance of time affects the solution of the algorithm since there are redundancies in picking the satisfaction instance ($\mathbf{x}(t)\models \mathcal{F}_{[a,b]}(g^{(1)}(\mathbf{x})\leq \epsilon_1)$ if $h^{(1)}(\mathbf{x}):=g^{(1)}(\mathbf{x})- \epsilon_1\leq  0$ at any single instance in $[a,b]$). By resetting these operators, the algorithm aims to locate a solution under feasible instances.
%%%%%%%%%%%%%%%%%%%%%%%%%%%%%%%%%%%%%%%%%%%%%%%%%%%%%%%%%%%%%%%%%%%%%%%%%%%%%%%%

\subsubsection{\texttt{GradientDescent}:}

\begin{function}
    \SetKwFunction{GradientDescent}{GradientDescent}
    \SetKwFunction{VD}{ValidityDomain}
    \SetKwFunction{SV}{SatisfactionVariable}
    \SetKwFunction{GC}{GradientComputation}
    \SetKwFunction{DO}{DistributedOptimisation}
    \caption{GradientDescent ()}    
    \label{tro:alg:GradientDescent}
    \KwIn{$z^i_{\text{inter}} = \{t^0, \mathbf{x}^i_{\text{inter}}\}, \text{step size}\ \delta,$ $\text{maximum iterations}\ L^\prime,$ \text{stopping criterion}$\ \eta$ }
    \KwOut{$z^i_{\text{opt}},\tau$}
    Receive neighbour states $\mathbf{x}^m_{\text{neigh}}$ for all $m\in\mathcal{N}_i$\;\label{tro:line:communication1}
    \ForAll{$\bar{\varphi}$ in $\varphi$}{
    $\mathrm{vd}_{ij}(\bar{\varphi})$ $\gets$ \VD($\bar{\varphi}$)\;}\label{tro:line:gd_vd}
    $k\gets 0$\;
    $\lambda_{ij}=0,\ \forall j$\;
    \Case{$t^0\in  \mathrm{vd}_{ij}^F(\bar{\varphi})$\label{tro:line:to_eventually}}
    {$\lambda_{ij}=1$\;}
    \Case{$t^0\in  \mathrm{vd}_{ij}^G(\bar{\varphi})$}
    {$\lambda_{ij}=1$\;}
    \Case{$t^0\in \bigcap_k \mathrm{vd}^F_{ik}(\bar{\varphi})$}
    {$\begin{cases}
        \lambda_{ij}&=1\ \text{for any one $j=k$}\\
        \lambda_{ij}&=0\ \text{otherwise}
    \end{cases}$}
    \Case{$t^0\in \bigcap_k\mathrm{vd}^G_{ik}(\bar{\varphi})$}
    {$\lambda_{ik}=1$\ for all $k$\;}\label{tro:line:to_always}
    $\mathbf{x}^i_{\text{opt}}\gets$ \DO{$\mathbf{x}^i$, $\delta$, $L'$, $\lambda_{ij}$}\;
    %\textcolor{violet}{placeholder}
    $z^i_{\text{opt}}=\{t^0,\mathbf{x}^i_{\text{opt}}\}$\;
    \ForAll{$\bar{\varphi}$ in $\varphi$\label{line:forall1}}
    {
    \uIf{$t^0\in\mathrm{vd}_{ij}(\bar{\varphi})$}{
    \uIf{$F^i(\mathbf{x}^i_{\text{opt}})\leq 0$\label{line:desiredminimum}}{
    \textbf{node} = \textbf{leaf}($\bar{\varphi}$)\;\label{line:leaf}
    $\tau(\textbf{leaf})=+1$\;\label{line:leaf+1}
    \While{\textbf{node} $\neq$ \textbf{root}($\bar{\varphi}$) and $\tau(\textbf{node})=+1$}
    {
    \textbf{node} = \textbf{parent}(\textbf{node})\;
    $\tau(\textbf{node}), t^\star\gets$ \SV(\textbf{node},$z^i_{\text{opt}}$)\;\label{line:root}
    }}
    \lElse{reset $\tau(\bar{\varphi}),t^\star$\label{line:endforall1}}
    %$\mathrm{vd}_{ij}(\bar{\varphi})$'s $\gets$ \VD($\bar{\varphi}, t^\star, t^0$)\;
    }
    }
    \Return $z^i_{\text{opt}}, \tau(\varphi)$\label{line:reset}
\end{function}
The function is presented in Function \ref{tro:alg:GradientDescent} and
computes the optimal value, $z^i_{\text{opt}}$, by solving the problem presented in equation \eqref{tro:eq:optimisation}. This allows the robots to compute vertices that locally do not violate the STL formula. 
 Once $z^i_{\text{opt}}$ is determined through Function \ref{tro:alg:optimisation}, the satisfaction variables are updated in Function \ref{tro:func:satisfactionvariable}. 

Based on the validity domain, the Function \ref{tro:alg:GradientDescent} determines which predicate functions are active in \eqref{optimisation_func} at every sampled time instance $t^0$. The Function \texttt{ValidityDomain}() in line \ref{tro:line:gd_vd} calculates the validity domains of each path $\bar{\varphi}$ based on Definition \ref{defn:validity_domain}. Let $K_i$ be the total number of independent and coupled predicate functions associated with robot $i$, a binary variable $\lambda_{ij}\in\{0,1\}, j\in\{1,\dots,K_i\}$ is assigned to determine whether a predicate function is active or not. It is set to 1 if the predicate is active and 0 otherwise. For example, 
\begin{itemize}
    \item If $\varphi_1 = \mathcal{G}_{[5,10]}(\|x_1-x_2\|\leq 2)$, then $\lambda_{11}=\lambda_{21}=1$ whenever $t^0\in[5,10]$ and 0 otherwise.
    \item If $\varphi_2 = \mathcal{F}_{[10,15]}(\|x_3\|\leq 5)$, then $\lambda_{31}=1$ whenever $t^0\in[10,15]$ and 0 otherwise. Once $x_3(t)\models \varphi_2$, $\lambda_{31}=0\ \forall t$.
\end{itemize}
The indices $i$ and $j$ in $\lambda_{ij}$ and $\mathrm{vd}_{ij}$ refer to robot $i$ and the $j$th predicate function associated with robot $i$, respectively. Here $j\in\{1,\dots,K_i\}$. 
We distinguish three cases: if the sampled point belongs to the validity domain of a single \emph{eventually} operator and/or a single \emph{always} operator, $\lambda_{ij}=1$. If the sampled point belongs to the validity domain of multiple \emph{eventually} operators, we activate only one of them at random, that is, $\lambda_{ij}=1$ only for one of them. This avoids enforcing conflicting predicates as it can happen that multiple \emph{eventually} operators may not be satisfied at the same time instance (For example $\varphi = \mathcal{F}_{[0,1]}(x>0)\land\mathcal{F}_{[0,1]}(x<0)$); see lines \ref{tro:line:to_eventually}-\ref{tro:line:to_always}.

In lines \ref{line:forall1}-\ref{line:endforall1}, the algorithm updates the satisfaction variable of all paths in the STL formula that impose restrictions on robot $i$'s states. The algorithm goes bottom-up, starting from the \textbf{leaf} node to the \textbf{root} node. First, it determines if $z_{opt}^i$ is the desired minimum (i.e., $F^i(\mathbf{x}_{opt}^i)\leq0$) in line \ref{line:desiredminimum}, and in lines \ref{line:leaf}-\ref{line:root}, the algorithm updates the satisfaction variable of all nodes in the path $\bar{\varphi}$ through the function \texttt{SatisfactionVariable()}.  If $z_{\text{opt}}^i$ is not the desired minimum, then all the satisfaction variables of the path $\bar{\varphi}$ are reset to $-1$ in line \ref{line:reset}. This could result from conflicting predicates at the same time instance.

%%%%%%%%%%%%%%%%%%%%%%%%%%%%%%%%%%%%%%%%%%%%%%%%%%%%%%%%%%%%%%%%%%%%%%%%%%%%%%%%

\subsubsection{\texttt{SatisfactionVariable}:}

\begin{function}
    \SetKwFunction{SV}{SatisfactionVariable}
    \caption{SatisfactionVariable()}
    \label{tro:func:satisfactionvariable}
    \KwIn{$\textbf{node},\ z^i_{\text{opt}}=\{t^0,\mathbf{x}^i\}$}
    \KwOut{$\tau,t^\star$}
    %\uCase{\textbf{leaf}$(\bar{\varphi})$}
    %{\lIf{$\mu^{h_{ij}}=\top $}{\Return $\tau=+1,t^\star$} \lElse{reset $\tau(\bar{\varphi}),t^\star$\ \Return $\tau,t^\star$}}
    \uCase{$(\textbf{node}=\mathcal{F}_I)$}{
    $\tau(\mathcal{F}_I)=+1$\;\label{line:eventually+1}
    %{$\mathcal{F}_I\neq \mathrm{\mathbf{root}}$}
    $t^\star=t^0$\;
    %\SV(\textbf{parent}($\mathcal{F}_I$), $z^i_{\text{opt}})$} 
    {\Return $\tau,t^\star$}\;}
    \uCase{$(\textbf{node}=\mathcal{G}_I)$}
    {\uIf{$\mathrm{\mathbf{robust}}$($\mathcal{G}_I$)$\ \geq 0$}{$\tau(\mathcal{G}_I)=+1$\;\label{line:always+1}
    }{\Return $\tau,t^\star$}}
    \uCase{$(\textbf{node}=\land)$}
    {$\tau(\land)=+1$\;\label{line:and+1}
    \Return $\tau,t^\star$}
\end{function}

This function, presented in Function \ref{tro:func:satisfactionvariable}, updates the \emph{satisfaction variable tree}, $\tau$. The aforementioned procedure decides if the satisfaction variable corresponding to each node listed is $+1$ (satisfied) or $-1$ (not yet satisfied). The discussion of handling disjunction operators is deferred to Section \ref{subsec:disjunctions}, as they are handled differently.  Considering the premise that the predicate is true, as indicated in line \ref{line:desiredminimum} of Function \ref{tro:alg:GradientDescent}, we evaluate the satisfaction variable as follows:

\begin{itemize}
    \item $\mathcal{F}_I$: The satisfaction variable of the eventually operator is updated along with $t^\star=t^0$.  This updated $t^\star$ is used to determine the new validity domains in line \ref{tro:line:gd_vd} of Function \ref{tro:alg:GradientDescent}; see Example 3 for an illustration of this procedure.
    \item $\mathcal{G}_I$: Unlike the \emph{eventually} operator, determining $\tau(\mathcal{G}_I)$ necessitates the computation of robustness over the entire validity domain of the operator. The function $\mathbf{robust()}$ uses the robust semantics of the STL presented in \cite{maler2004monitoring}. Particularly, it samples a user-defined number of points in the interval $\mathrm{vd}^G_{ij}()$ and computes $\inf_{t\in\mathrm{vd}_{ij}^G} h^d_{i,l}(\mathbf{x}^i(t))$ or $\inf_{t\in\mathrm{vd}_{ij}^G} h^c_{i,j}(\mathbf{x}^i(t))$. If the robustness is non-negative, indicating satisfaction of the task, the value of $\tau(\mathcal{G}_I)$ is updated to $+1$. 
    \item $\land$: This set node returns the satisfaction variable as $+1$ since it does not impose spatial or temporal restrictions.
\end{itemize}

%%%%%%%%%%%%%%%%%%%%%%%%%%%%%%%%%%%%%%%%%%%%%%%%%%%%%%%%%%%%%%%%%%%%%%%%%%%%%%%%

\subsubsection{Branch-and-Pick for Disjunctions:}\label{subsec:disjunctions}

\begin{figure}
     \centering
     \includegraphics[width=0.99\columnwidth]{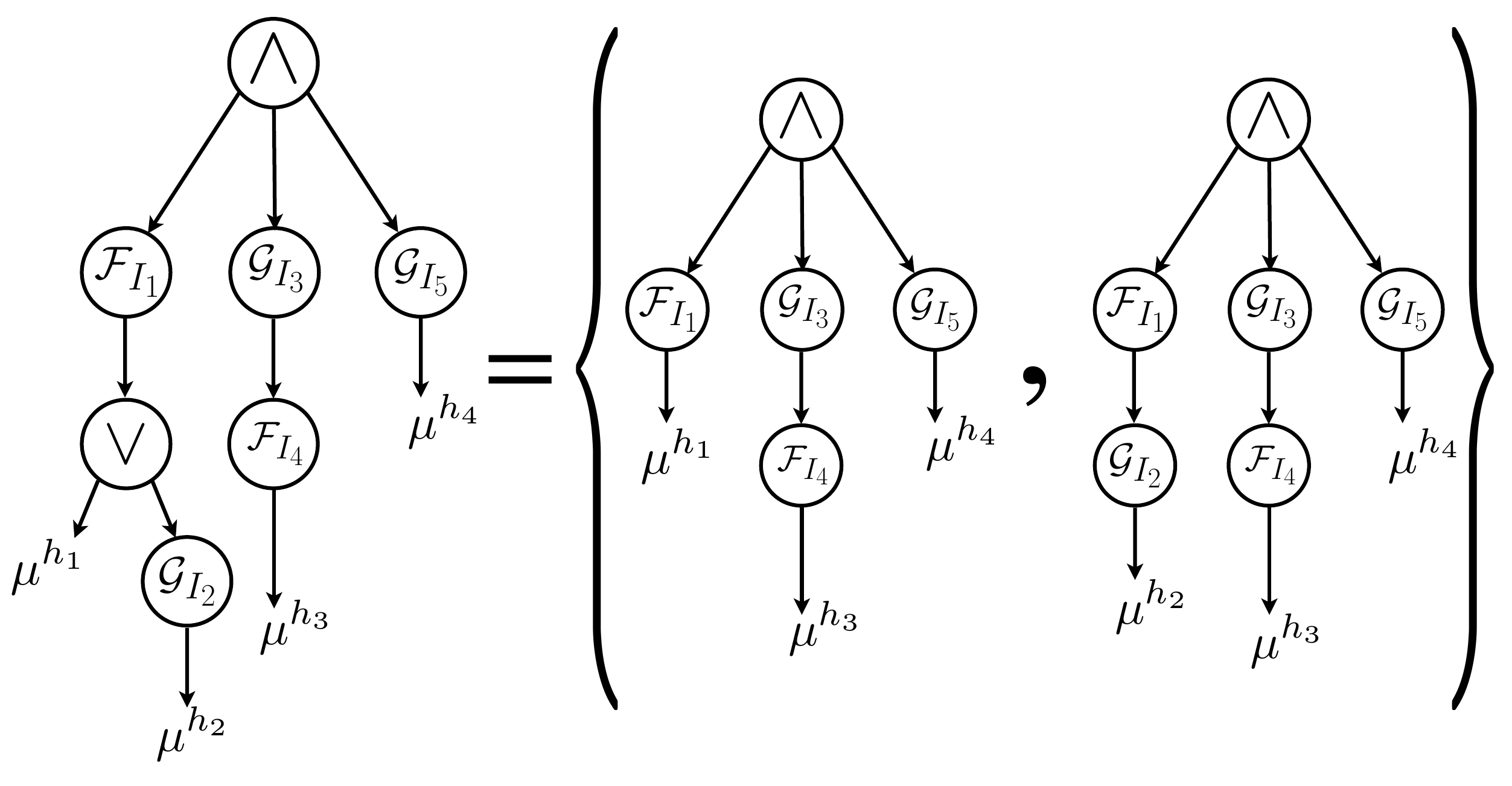}
     \caption{Disjunction representation for disjunctive components using STL parse tree.}
     \label{fig:disjunction}
\end{figure}

In our approach, we address disjunctions as follows: Given an STL formula of the form $\varphi = \bigvee_{i\in {1,\dots,K}} \phi_i$, which can also be represented as $\varphi = \lor (\phi_1,\phi_2,\dots,\phi_K)$, we divide it into $K$ individual STL formulas. The agents then run Algorithm \ref{tro:alg:maps} separately for each $\varphi = \phi_i$, where $i \in {1,\dots,K}$. For instance, consider the STL formula represented as \eqref{eq:stl_eg}, 
\begin{align*}
    \varphi &= \mathcal{F}_{I_1}\Big({\mu^{h_1}}\lor \mathcal{G}_{I_2} ({\mu^{h_2}})\Big) \land \mathcal{G}_{I_3}\mathcal{F}_{I_4}({\mu^{h_3}})\land \mathcal{G}_{I_5}({\mu^{h_4}}).
\end{align*}
We branch it into two STL formulas: $\phi_1=\mathcal{F}_{I_1}{\mu_1} \land \mathcal{G}_{I_3}\mathcal{F}_{I_4}({\mu_3})\land \mathcal{G}_{I_5}({\mu_4})$ and $\phi_2=\mathcal{F}_{I_1}\mathcal{G}_{I_2} ({\mu_2}) \land \mathcal{G}_{I_3}\mathcal{F}_{I_4}({\mu_3})\land \mathcal{G}_{I_5}({\mu_4})$, as illustrated in Figure \ref{fig:disjunction}. The search terminates when any branch of the disjunction satisfies the condition $\tau(root)\neq +1$, as specified on line \ref{tro:main_for} of Algorithm 1. {
{We acknowledge that this naive method of handling disjunctions can result in exponential growth with the addition of more operators. An alternative approach, akin to the branch-and-bound method from optimisation \cite{morrison2016branch}, involves evaluating the robustness of each $\phi_i$ for $i\in{1,\dots,K}$ and executing \texttt{MAPS}$^2$ only for the formulas that show a faster increase in satisfaction. However, this strategy might necessitate a higher level of communication among robots which goes beyond their existing communication network and possibly require a central authority to coordinate task fulfilment. For example, the STL formula
\[
\varphi = \mathcal{G}_{[0,5]}(x_1<5)\lor (\mathcal{F}_{[0,5]}(|x_2-x_3|>2)).
\]
comprises disjunction between $\phi_1 = \mathcal{G}_{[0,5]}(x_1<5)$ and $\phi_2 = \mathcal{F}_{[0,5]}(|x_2-x_3|>2)$. Observe that $\phi_1$ requires no inter-robot communication, while $\phi_2$ necessitates communication between robots $2$ and $3$. In the implementation of a method akin to branch-and-bound, we would branch into two formulas, $\phi_1$ and $\phi_2$, and repeatedly switch between them if we observe the robustness of one formula decaying faster compared to the other. This switching must be performed by a central authority that observes the decay in robustness. If the switching is decided among the robots, then robot $1$ of $\phi_1$ needs to communicate the robustness decay with the network of robots $2$ and $3$. This requires robot $1$ to establish communication with the network of robots $2$ and $3$ in order to decide which branch to grow, thereby necessitating communication links where none existed before. Without such a communication link, both $\phi_1$ and $\phi_2$ would need to be satisfied using the naive approach presented in our work. This motivates our choice to use the naive approach.
}}

%%%%%%%%%%%%%%%%%%%%%%%%%%%%%%%%%%%%%%%%%%%%%%%%%%%%%%%%%%%%%%%%%%%%%%%%%%%%%%%%

%%%%%%%%%%%%%%%%%%%%%%%%%%%%%%%%%%%%%%%%%%%%%%%%%%%%%%%%%%%%%%%%%%%%%%%%%%
\subsection{Analysis}\label{tro:subsec:analysis}
In this section, we analyse the proposed algorithm and arrive at proving the probabilistic completeness. 

Let the set $\mathcal{S}\subseteq\mathcal{W}$ be a compact set where a trajectory $\mathbf{y}:[0,\mathrm{th}(\varphi)]\to \mathcal{S}$ satisfies the STL formula.  Along the lines of \cite{8584061}, let a trajectory $\mathbf{y}$ be located on the boundary of the set $\mathcal{S}$, the satisfiable set, dividing $\mathcal{W}$ into a feasible set $\mathcal{S}$ and an infeasible set $\mathcal{W}\backslash \mathcal{S}$.

Starting with an initial linear trajectory in the augmented time-space domain, each uniformly sampled time point $t^0$ corresponds to a position $\mathbf{x}_{\text{inter}}$ either in $\mathcal{S}$ or $\mathcal{W}\backslash \mathcal{S}$. If $\mathbf{x}_{\text{inter}}\in \mathcal{S}$, we leave it unchanged as it meets the requirements. But if $\mathbf{x}_{\text{inter}}\notin \mathcal{S}$, we use gradient descent to reach a point on $\mathbf{y}$, since it lies on the boundary of the constraints' set. 

\begin{figure}
    \centering
    \includegraphics[width=0.8\columnwidth]{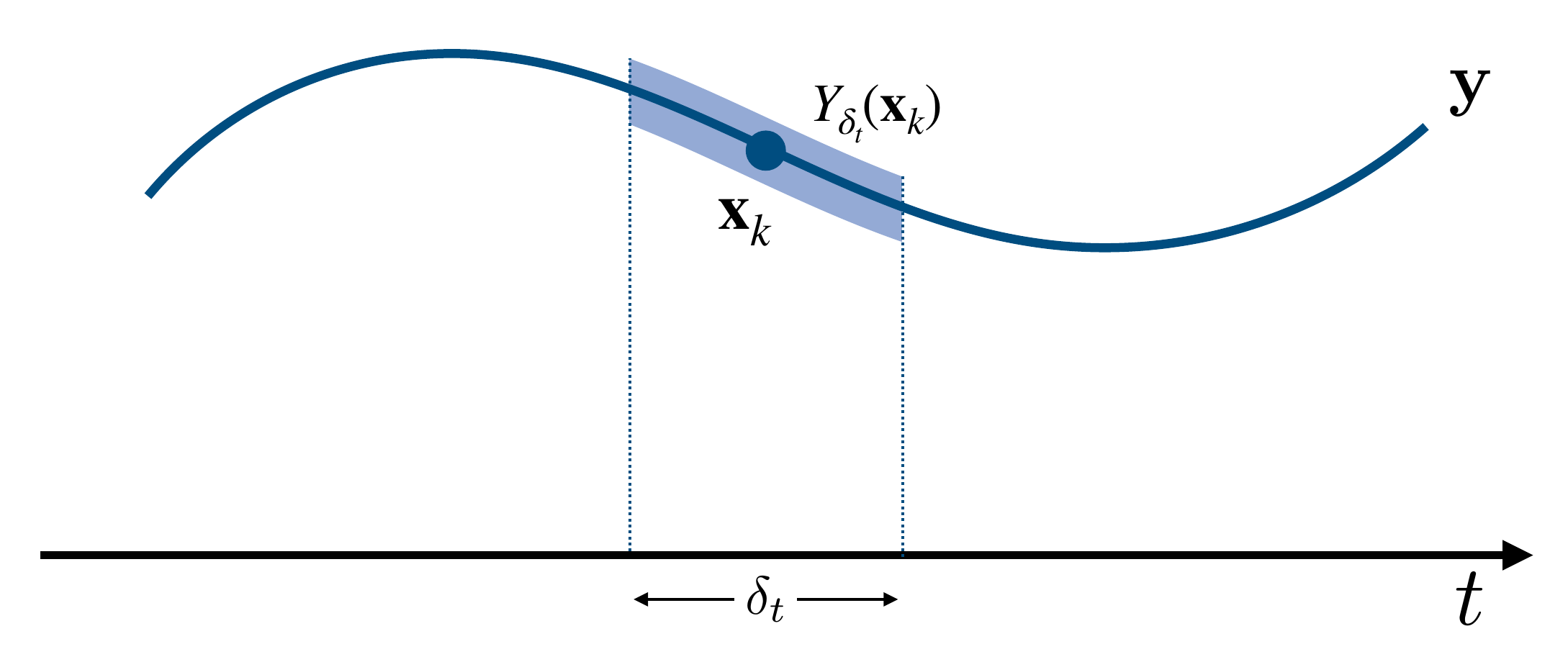}
    \caption{Illustration of $Y_{\delta_t}(\mathbf{x}_k)$.}
    \label{fig:completeness}
\end{figure}

Next, divide the trajectory $\mathbf{y}:[0,\mathrm{th}(\varphi)]\to\mathcal{S}$ into $L+1$ points $\mathbf{x}_k$, where $0\leq k \leq L$ and $\mathbf{y}(\mathrm{th}(\varphi))=\mathbf{x}_f=\mathbf{x}_L$ by dividing the time duration into equal intervals of $\delta_t$. Without loss of generality, assume that the points $\mathbf{x}_k$ and $\mathbf{x}_{k+1}$ are separated by $\delta_t$ in time. With $L\delta_t=\mathrm{th}(\varphi)$, the probability of sampling a point in an interval of length $\delta_t$ can be calculated as $p = \frac{\delta_t}{\mathrm{th}(\varphi)}$. If $\delta_t<<\mathrm{th}(\varphi)$, then $p<1/2.$
Denote the sequential covering class\footnote{Meaning $\mathbf{y} \subset \bigcup_{k=1}^{L} Y_{\delta_t}(\mathbf{x}_k)$} of trajectory $\mathbf{y}$ as $Y_{\delta_t}(\mathbf{x}_k)$. The length of $Y_{\delta_t}(\mathbf{x}_k)$ is $\delta_t$ in the time domain and is centered at $\mathbf{x}_k$. See Figure \ref{fig:completeness} for reference. 
A trial is counted as successful if we sample a point $t^0$ within the interval $\delta_t/2$ on either side of $\mathbf{x}_k$, that is, within $Y_{\delta_t}(\mathbf{x}_k)$. If there are $L$ successful trials, the entire trajectory $\mathbf{y}$ is covered, and the motion planning problem is solved. Consider $m$ total samples, where $m\gg L$, and treat this as $m$ Bernoulli trials with success probability $p$ since each sample is independent with only two outcomes. We are now ready to state the following lemma. 
\begin{lemma}\label{lemma:covering}
Let a constant $L$ and probability $p$ such that $p < \frac{1}{2}$. Further, let $m$ represent the number of samples taken by the \texttt{MAPS}$^2$ algorithm. Then, the probability that \texttt{MAPS}$^2$ fails to sample a segment after $m$ samples is at most $\frac{(m-L)p}{(mp-L)^2}$.
\end{lemma}

\begin{proof}
The probability of not having $L$ successful trials after $m$ samples can be expressed as:
\[\mathbf{P}[X_m\leq L]=\sum_{k=0}^{L-1} \binom{m}{k}p^k(1-p)^{m-k}\]
and according to \cite{feller1}, if $p<\frac{1}{2}$, we can upper bound this probability as:
\[\mathbf{P}[X_m\leq L]\leq \frac{(m-L)p}{(mp-L)^2}.\]
As $p$ and $L$ are fixed and independent of $m$, the expression $\frac{(m-L)p}{(mp-L)^2}$ approaches 0 with as $m$ increases, thus completing the proof. 
\end{proof}
Next, we present a final lemma which helps us prove the probabilistic completeness of the algorithm. 
{
%\begin{lemma}
%The probability that any valid trajectory segment remains uncovered by the algorithm asymptotically approaches zero as the number of iterations increases.
%\end{lemma}
{
\begin{lemma}\label{lemma:label}
No sampled point $\mathbf{x}_k$ is falsely labelled as satisfying the STL formula $\varphi$ unless it actually does.
\end{lemma}
\begin{proof}
The algorithm initiates by setting all satisfaction variables, $\tau$, to ${-1}$, as inputs to Algorithm \ref{tro:alg:maps}. These variables are updated in Function \ref{tro:func:satisfactionvariable} designed for evaluating whether $\tau$ meets the satisfaction criteria. The function adjusts $\tau$ in accordance with the definition of STL operators presented in Section \ref{subsec:stl}, ensuring that updates accurately reflect the satisfaction status. Furthermore, the update to $\tau(\textbf{leaf})$ within Function \ref{tro:alg:GradientDescent} (referenced at line \ref{line:leaf+1}) occurs only when the condition $F^i\leq 0$ is met. This condition indicates that all active predicates are satisfied by definition. Thus, no satisfaction variable is incorrectly updated. 
\end{proof}
}
Next, the paper's final result is presented, which states that the probability of the algorithm providing an STL formula satisfying trajectory (if one exists) approaches one as the number of samples tends to infinity. This is a desirable property for sampling-based planners and such algorithms are termed probabilistically complete.
{
\begin{thm}
Algorithm \ref{tro:alg:maps} is probabilistically complete.
\end{thm}
\begin{proof}
The proof follows from Lemmas \ref{lemma:constraints}, \ref{lemma:covering}, and \ref{lemma:label}. 
From Lemma \ref{lemma:constraints} and Lemma \ref{lemma:label}, we know that every sample added to the trajectory satisfies the STL formula. Thus, what needs to be shown is that the algorithm samples infinitely many times and covers the entire time horizon. From Lemma \ref{lemma:covering}, we know that the probability of covering the entire time horizon is $1-\mathbf{P}[X_m\leq L]$. Suppose the Algorithm \ref{tro:alg:maps} reaches $J=L'$ samples without finding a feasible solution, then it discards $J$ samples as seen in line \ref{line:maps:tauF} of Algorithm \ref{tro:alg:maps}. Given Assumption \ref{ass:feasible}, we have $J<\infty$, and since $J$ is the number of discarded samples, we also have $J\leq m$ where $m$ is the total number of samples sampled so far (including the discarded ones). Thus, the probability of the trajectory satisfying the STL formula is $1-\frac{((m-J)-L)p}{((m-J)p-L)^2}$, which approaches one as $m\to\infty$. Thus, the algorithm is probabilistically complete.
\end{proof}
}

\begin{rmrk}\label{rmrk:smooth}
    Our algorithm can be endowed in a post-processing stage with a module that smoothens the trajectory to avoid large accelerations. However, care needs to be taken since the smoothened paths may no longer satisfy the STL formula. One could also use more sophisticated approaches like B-splines to impose velocity and acceleration limits as shown in \cite{10528798}. 
\end{rmrk}
\begin{rmrk}\label{rmrk:dynamic}
At present, our approach does not incorporate kinematic or dynamic constraints. Incorporation of such constraints could be attempted by either deploying the kinodynamic version of the RRT algorithm \cite{Webb2012KinodynamicRO}, or by using an existing low-level controller to track the generated open-loop trajectories. Some examples of such controllers include the Model Predictive Controller \cite{862901} and the input constrained Prescribed Performance Controller \cite{10273607, 10138653}. 
This incorporation is by no means straightforward but requires fusion with another type of methodological machinery that goes beyond the scope of the current work. Moreover, such controllers have been developed for a large variety of dynamical systems and hence the proposed algorithm is practical and applicable to a large class of robots. %We believe this approach strikes a balance between theoretical guarantees and practical feasibility.
\end{rmrk}

%%%%%%%%%%%%%%%%%%%%%%%%%%%%%%%%%%%%%%%%%%%%%%%%%%%%%%%%%%%%%%%%%%%%%%%%%%%%%%%%

\section{Simulations}\label{sec:simulations}

In this section, we present simulations of various scenarios encountered in a multi-robot system. Restrictions are imposed using an STL formula and \text{MAPS}$^2$ is utilised to create trajectories that comply with the STL formula. In the following we consider 4 agents, with $\delta=0.1$, $\eta=0.01$ and $L=L'=100$. The simulations were run on an 8 core Intel\textsuperscript{\textregistered} Core\textsuperscript{\texttrademark} i7 1.9GHz CPU with 16GB RAM.  \endnote{The project code can be found here \url{https://github.com/sewlia/Maps2}}

\begin{figure*}
     \centering
     \begin{subfigure}[t]{0.245\textwidth}
         \centering
         \includegraphics[width=\textwidth]{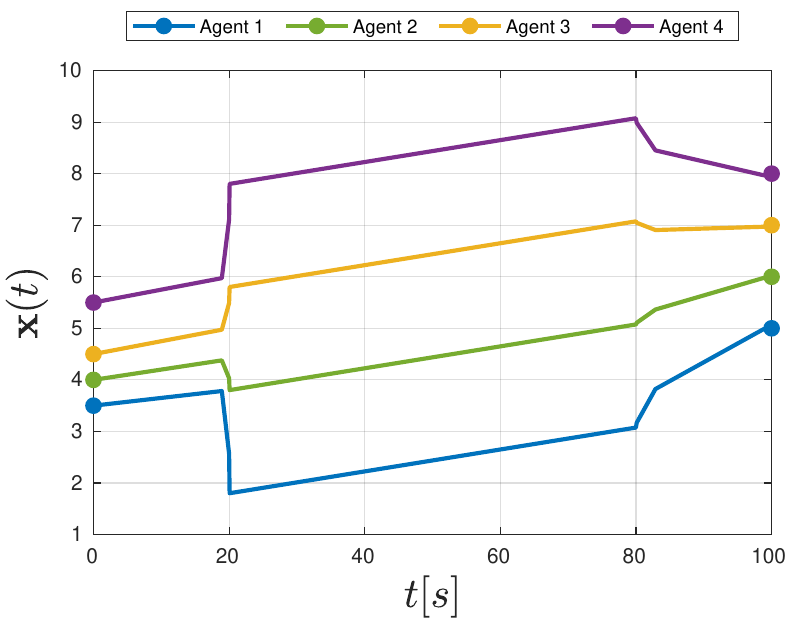}
         \subcaption{Collision avoidance.}
         \label{fig:sim:collision}
     \end{subfigure}%
     \begin{subfigure}[t]{0.245\textwidth}
         \centering
         \includegraphics[width=\textwidth]{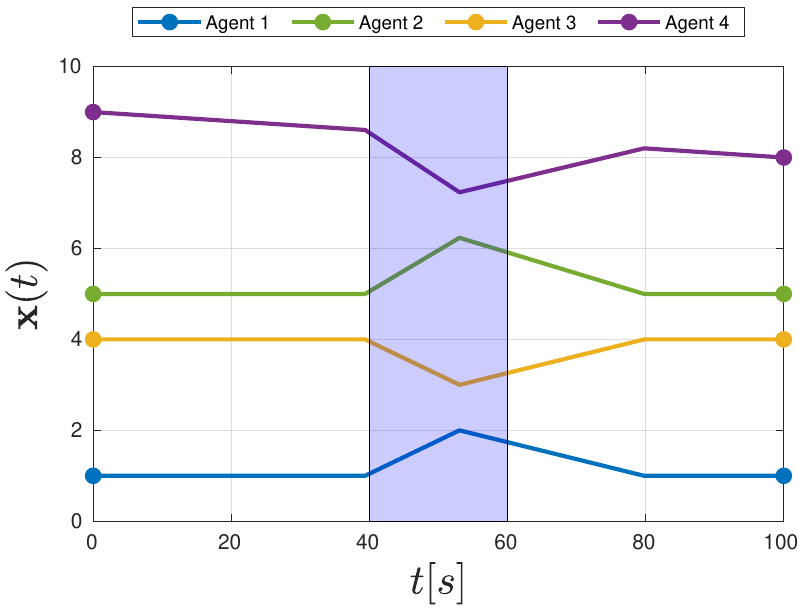}
         \subcaption{Rendezvous.}
         \label{fig:sim:rendezvous}
     \end{subfigure}%
     \begin{subfigure}[t]{0.245\textwidth}
         \centering
         \includegraphics[width=\textwidth]{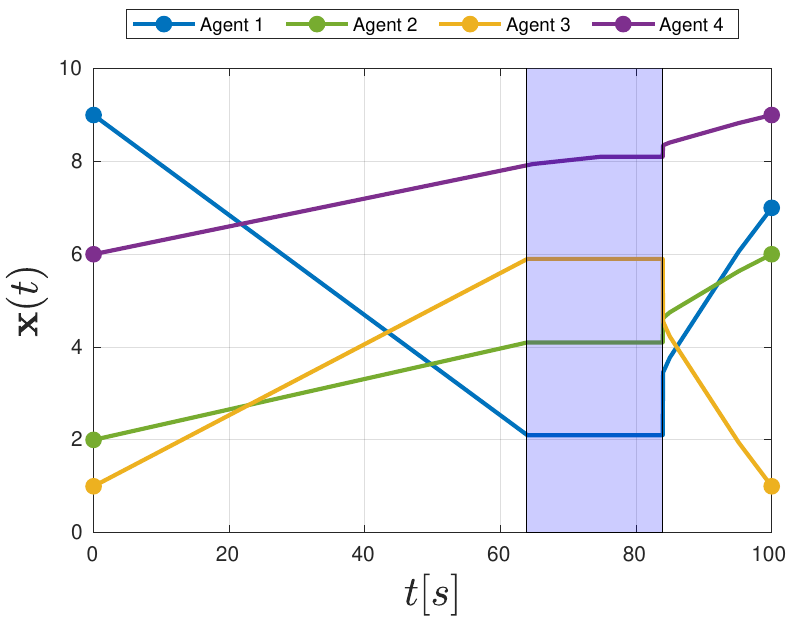}
         \subcaption{Stability.}
         \label{fig:sim:stability}
     \end{subfigure}
     \begin{subfigure}[t]{0.245\textwidth}
         \centering
         \includegraphics[width=\textwidth]{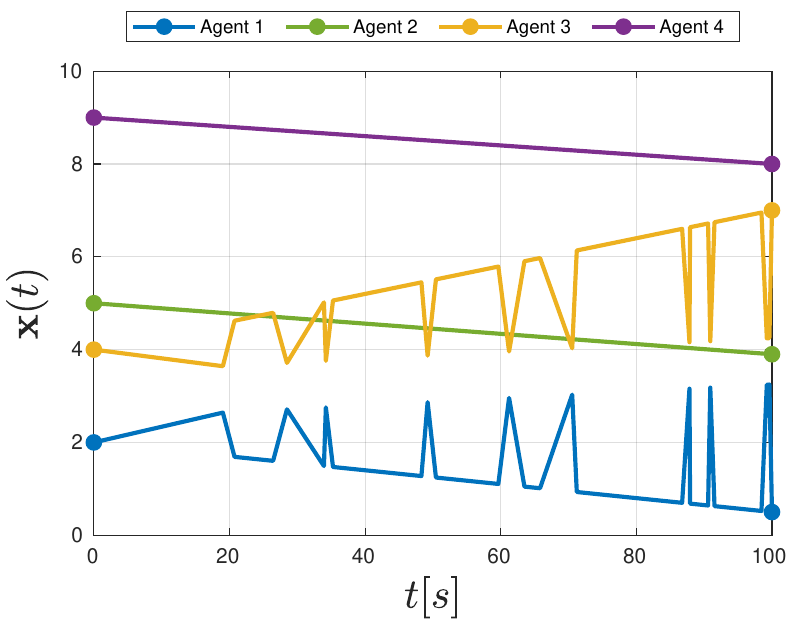}
         \subcaption{Recurring tasks.}
         \label{fig:sim:infinitelyoften}
     \end{subfigure}%
        \caption{Simulation results of MAPS$^2$ with four agents.}
        % \label{fig:experiment_plots}
        %\vspace{-0.4cm}
\end{figure*}

\subsubsection{Collision avoidance} We begin with a fundamental requirement in multi-robot systems: avoiding collisions. In this scenario, it is assumed that all agents can communicate or sense each other's positions. The following STL formula is used to ensure collision avoidance in the interval $20$[s] to $80$[s]:
\[
\varphi = \mathcal{G}_{[20,80]} (\|x_i-x_j\|\geq 1)
\]
where $\{i,j\}\in\{\{1,2\},\{1,3\},\{1,4\},\{2,3\},\{2,4\},\{3,4\}\}$. As depicted in Figure \ref{fig:sim:collision}, all four agents maintain a distance of at least 1 unit from each other during the interval $[20,80]$[s]. The maximum computation time by any agent is $0.1143$[s].

\subsubsection{Rendezvous} The next scenario is rendezvous. We use the eventually operator to express this requirement. The STL formula specifies that agents 1 and 3 must approach each other within 1 distance unit during the interval $[40,60]$[s] and similarly, agents 2 and 4 must meet at a minimum distance of 1 unit during the same interval. The STL formula is:
\[
\varphi = \mathcal{F}_{[40,60]} (\|x_1-x_3\|\leq 1 \land \|x_2-x_4\|\leq 1).
\]
As seen in Figure \ref{fig:sim:rendezvous}, agents 1 and 3 and agents 2 and 4 approach each other within a distance of 1 unit during the specified interval. It's worth noting that the algorithm randomly selects the specific time $t^\star$ within the continuous interval $[40,60]$[s] at which the satisfaction occurs. 
The maximum computation time by any agent is $0.0637$[s].

\subsubsection{Stability} The last task is that of stability, which is represented by the STL formula $\mathcal{F}_{[a_1,b_1]}\mathcal{G}_{[a_2,b_2]}(g^{(1)}(\mathbf{x})\leq \epsilon_1)$. This formula requires that $(g^{(1)}(\mathbf{x})\leq \epsilon_1)$ must always hold within the interval $[t^\star + a_2, t^\star + b_2]$, where $t^\star \in [a_1,b_1]$. This represents stability, as it requires $(g^{(1)}(\mathbf{x})\leq \epsilon_1)$ to always hold within the interval $[t^\star + a_2, t^\star + b_2]$, despite any transients that may occur in the interval $[a_1,t^\star)$. Figure \ref{fig:sim:stability} presents a simulation of the following STL formula:
\begin{align*}
\varphi =\ \mathcal{F}_{[0,100]}\ \mathcal{G}_{[0,20]}&\ \Big((1.9\leq x_1\leq 2.1)\land (3.9\leq x_2\leq 4.1)\\ &\land (5.9\leq x_3\leq 6.1)\land (7.9\leq x_4\leq 8.1)\Big)
\end{align*}
where $t^\star = 63.97$[s]. The maximum computation time by any agent is $0.0211$[s].

\subsubsection{Recurring tasks} 
The next scenario is that of recurring tasks. This can be useful when an autonomous vehicle needs to repeatedly survey an area at regular intervals, a bipedal robot needs to plan periodic foot movements, or a ground robot needs to visit a charging station at specified intervals. The STL formula to express such requirements is given by $\mathcal{G}_{[a_1,b_1]}\mathcal{F}_{[a_2,b_2]}(g^{(1)}(\mathbf{x})\leq \epsilon_1)$, which reads as `beginning at $a_1$[s], $g^{(1)}(\mathbf{x})\leq \epsilon_1$ must be satisfied at some point in the interval $[a_1+a_2, a_1+b_2]$[s] and this should be repeated every $[b_2-a_2]$[s].' A simulation of the following task is shown in Figure \ref{fig:sim:infinitelyoften}:
\[
\varphi = \mathcal{G}_{[0,100]}\mathcal{F}_{[0,20]} (\|x_1-x_3\|\leq 1).
\]
Every $20$[s], the condition $|x_1-x_3|\leq 1$ is met. It's worth noting that the specific time $t^\star$ at which satisfaction occurs is randomly chosen by the algorithm. The maximum computation time by any agent is $0.2017$[s].

In reference to Remark \ref{rmrk:smooth}, an example of post-processing the trajectories is shown in Figure \ref{fig:inf_paper} for the STL formula,
\begin{equation}\label{eq:inf_paper}\varphi = \mathcal{G}_{[0,100]}\mathcal{F}_{[0,20]} (\|x_1-x_2\|\leq 1).
\end{equation}
A 3rd order polynomial was applied using the Savitzky-Golay filter to smoothen the trajectory. Smoothening helps to avoid any large accelerations and sudden velocity changes, though it may come at the cost of potential STL violations.   
\begin{figure}
    \centering
    \includegraphics[width=0.9\columnwidth]{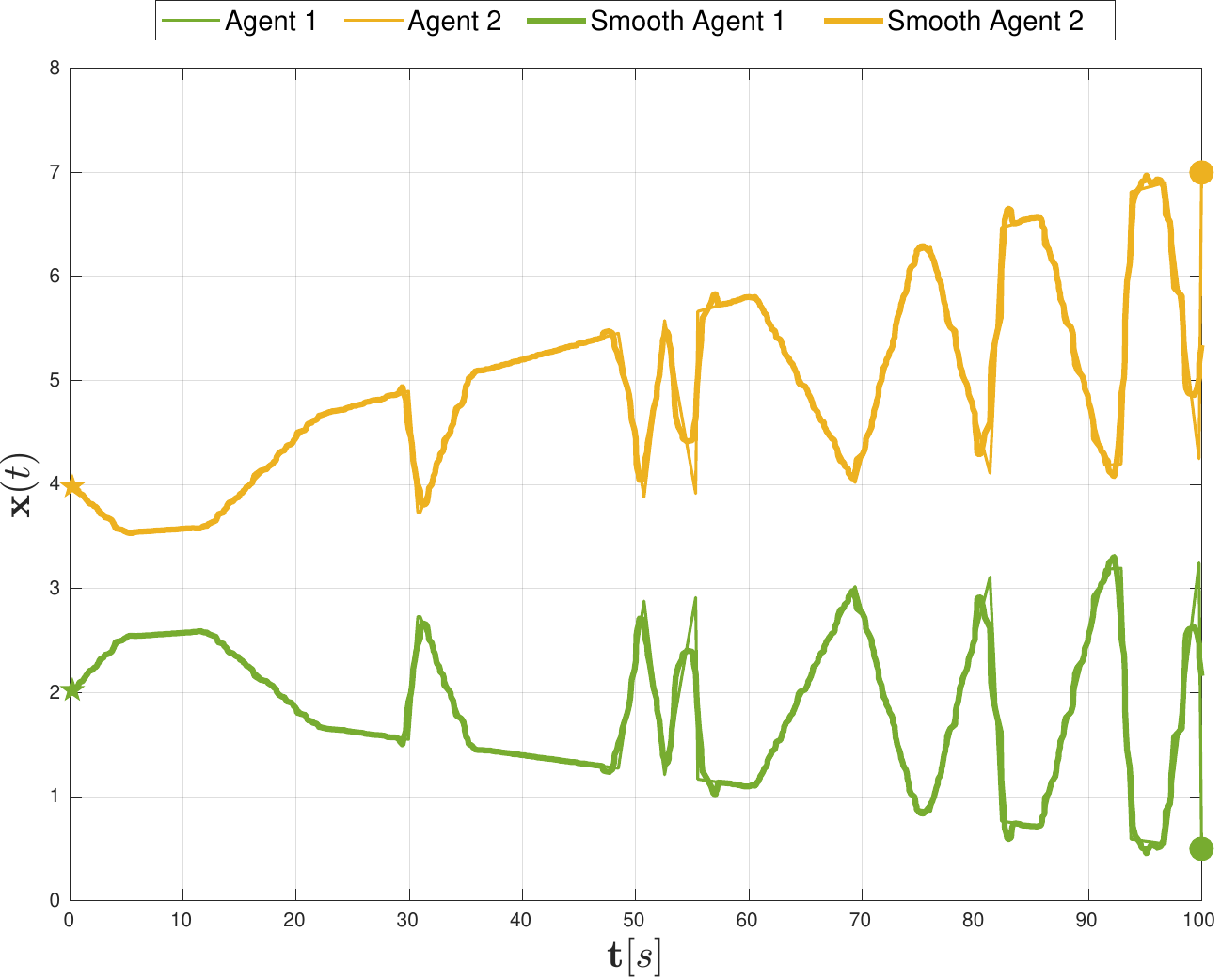}
    \caption{{Non-smooth and smooth paths for the formula \eqref{eq:inf_paper}}.}
    \label{fig:inf_paper}
\end{figure}

\begin{figure}
    \centering
    \includegraphics[width=\columnwidth]{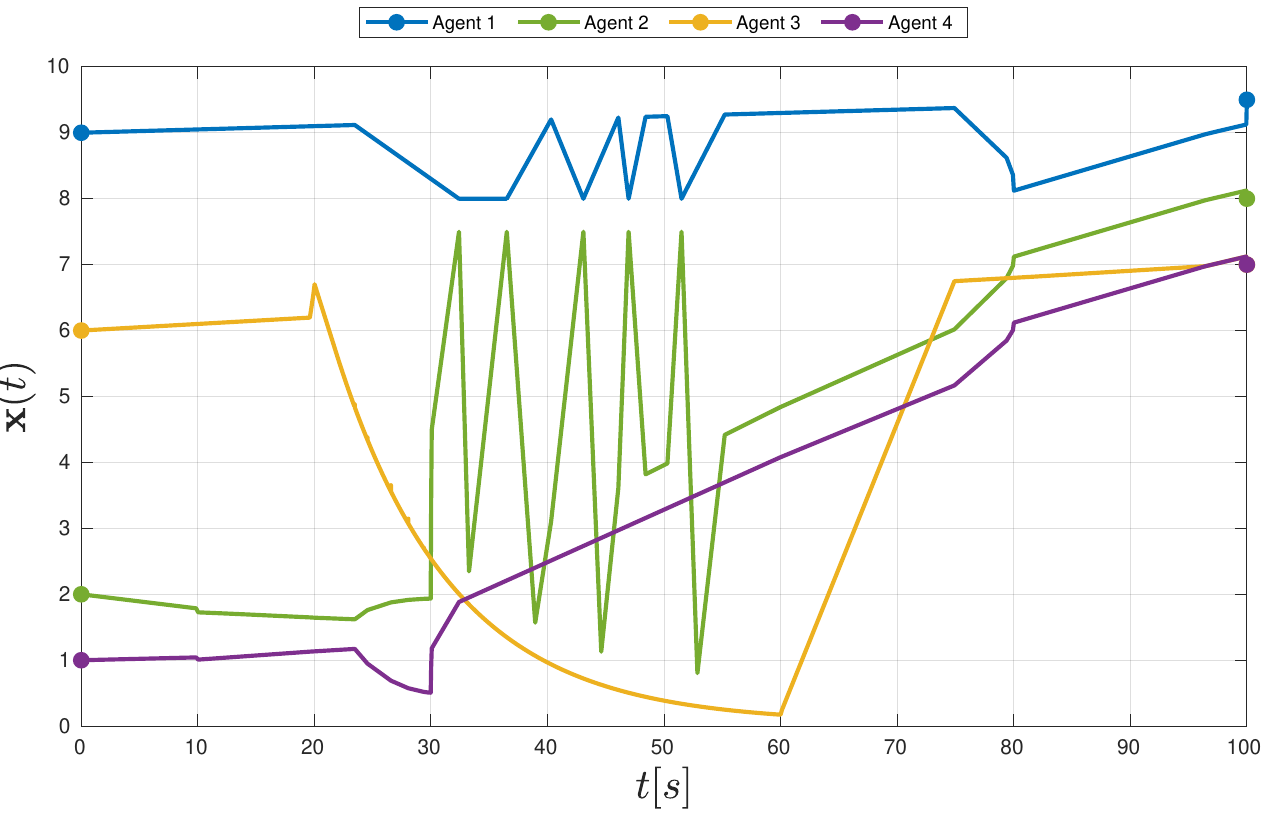}
    \caption{Overall case study.}
    \label{fig:sim:overall}
\end{figure}

{
{
\subsubsection{Multi-agent case study}
In this case study, we design trajectories for a team of 100 agents that exist in a $100\times 100$[m] space and $[0,100]$[s] time span. The team needs to adhere to the following STL formula,
\begin{equation}\label{eq:100agents}
\varphi = \mathcal{G}_{[10,90]}\Big[\|x_i-x_j\|\geq 0.01 \land \|x_i-(50,50)\|\leq 5\Big]
\end{equation}
$\forall i,j \in \{1,2,\dots,100\}$ and $i\neq j$. Note that the above STL formula has $5150$ predicates. In the interval $[10,90]$[s], the STL formula dictates every agent to be at least $0.01$[m] apart from every other agent and to be at least $5$[m] close to the centre point $(50,50)$[m]. The simulation results are shown in Figure \ref{fig:100agents} where the Figures \ref{fig:t0_overall}-\ref{fig:t0_xyview} are the trajectories before the start of the algorithm while Figures \ref{fig:tf_overall}-\ref{fig:tf_xyview}  shows the trajectories at the end of $j=1000$ iterations, as mentioned in Algorithm \ref{tro:alg:maps}. 
The simulation took $17.84$[s] to complete without parallelisation. The faster computation can be attributed to the nature of the design of the cost function in \eqref{tro:eq:optimisation}, which allows for points that do not violate the formula not to be changed. 
The robustness of the STL formula is shown in Figure \ref{fig:robust}, a negative robustness signifies task satisfaction. Here, the robustness converges to $0$, because the robustness for an \textit{always} operator reflects the worst-case scenario. It is important to note that computing the result for Figure \ref{fig:robust} required 12 hours and 10 minutes of computation time since it had to be performed centrally.}
}
\begin{figure*}
     \centering
     \begin{subfigure}[t]{0.325\textwidth}
         \centering
         \includegraphics[width=\textwidth]{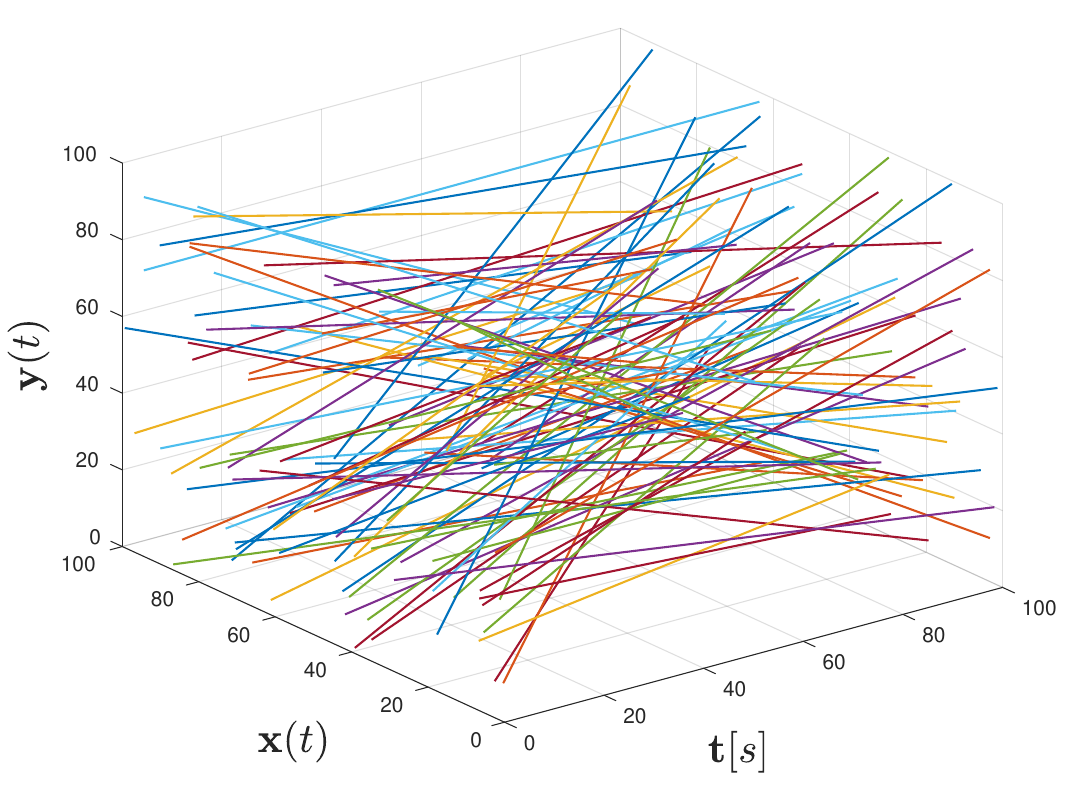}
         \subcaption{3d view at $j = 0$.}
         \label{fig:t0_overall}
     \end{subfigure}%
     \begin{subfigure}[t]{0.325\textwidth}
         \centering
         \includegraphics[width=\textwidth]{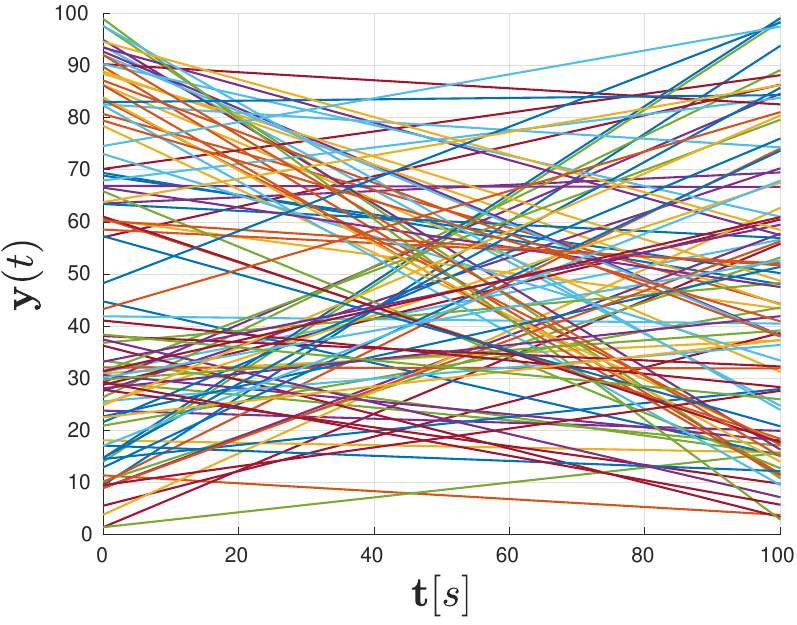}
         \subcaption{$t-y$ axis view at $j=0$.}
         \label{fig:t0_tyview}
     \end{subfigure}%
     \begin{subfigure}[t]{0.325\textwidth}
         \centering
         \includegraphics[width=\textwidth]{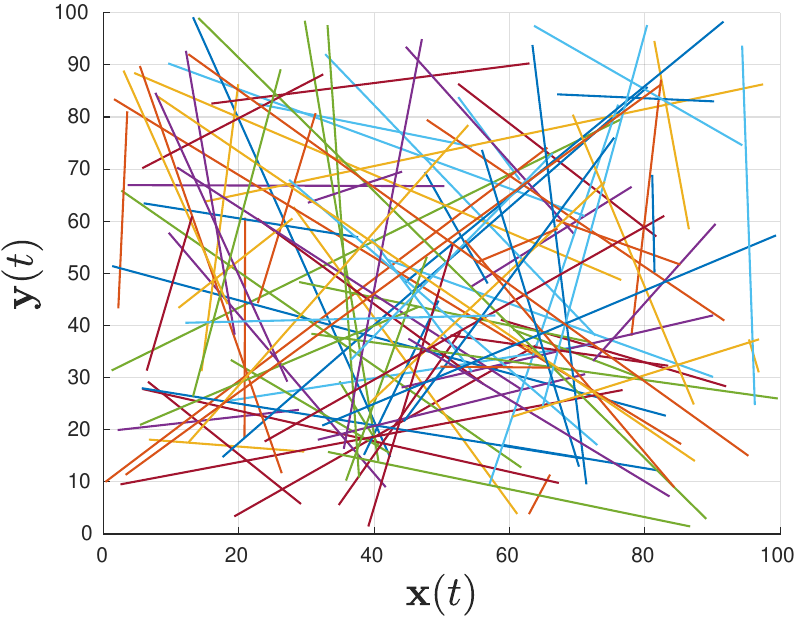}
         \subcaption{$x-t$ axis view at $j=0$.}
         \label{fig:t0_xyview}
     \end{subfigure}\\
     \begin{subfigure}[t]{0.325\textwidth}
         \centering
         \includegraphics[width=\textwidth]{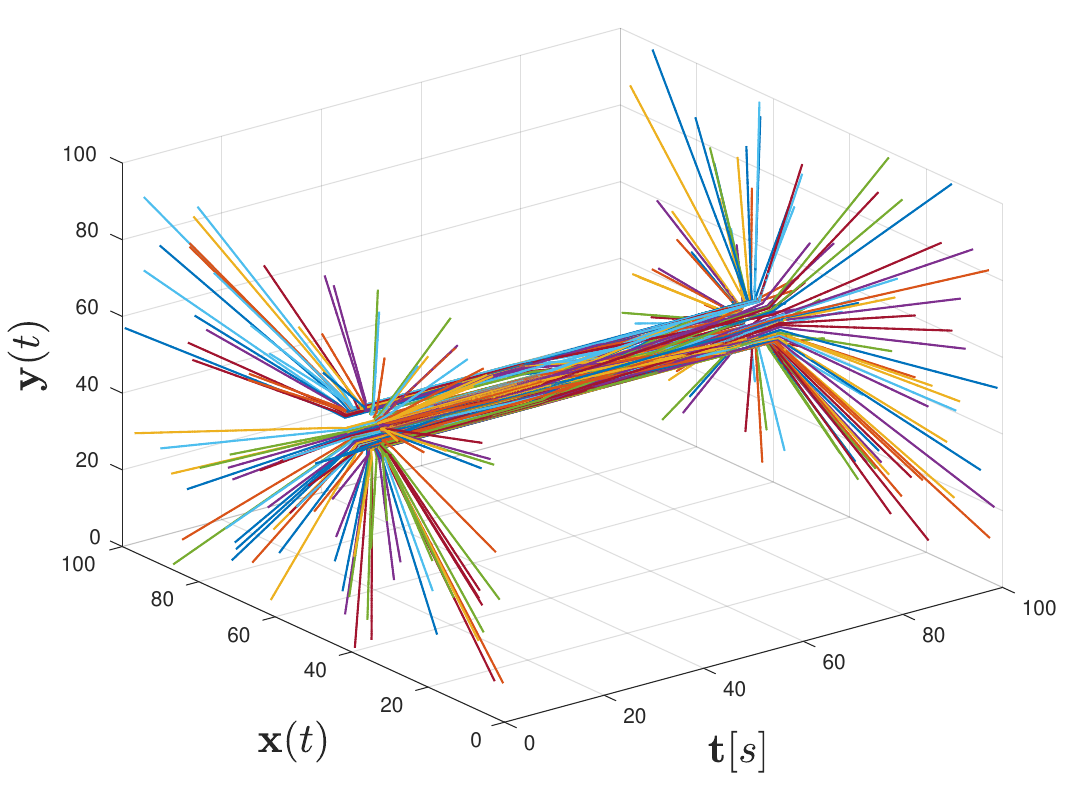}
         \subcaption{3d view when $j=1000$.}
         \label{fig:tf_overall}
     \end{subfigure}%
     \begin{subfigure}[t]{0.325\textwidth}
         \centering
         \includegraphics[width=\textwidth]{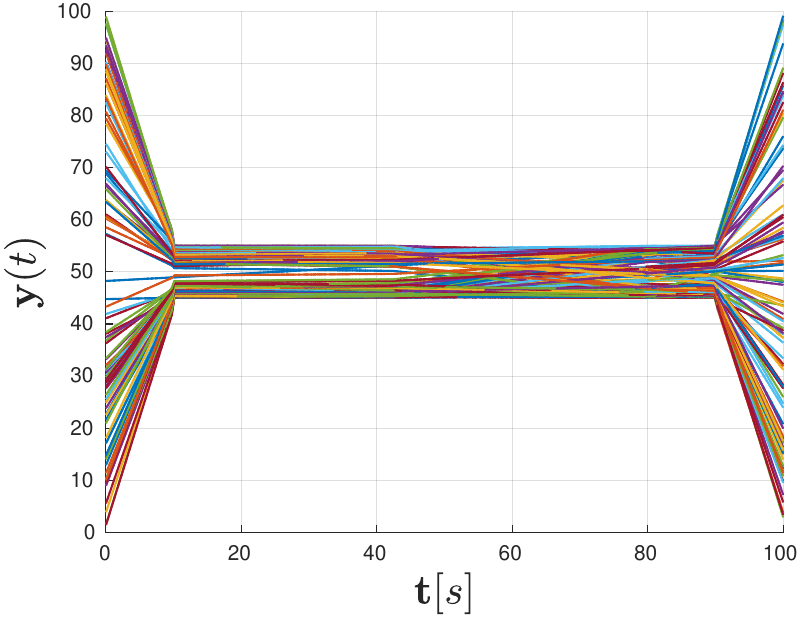}
         \subcaption{$t-y$ axis view at $j=1000$.}
         \label{fig:tf_tyview}
     \end{subfigure}%
     \begin{subfigure}[t]{0.325\textwidth}
         \centering
         \includegraphics[width=\textwidth]{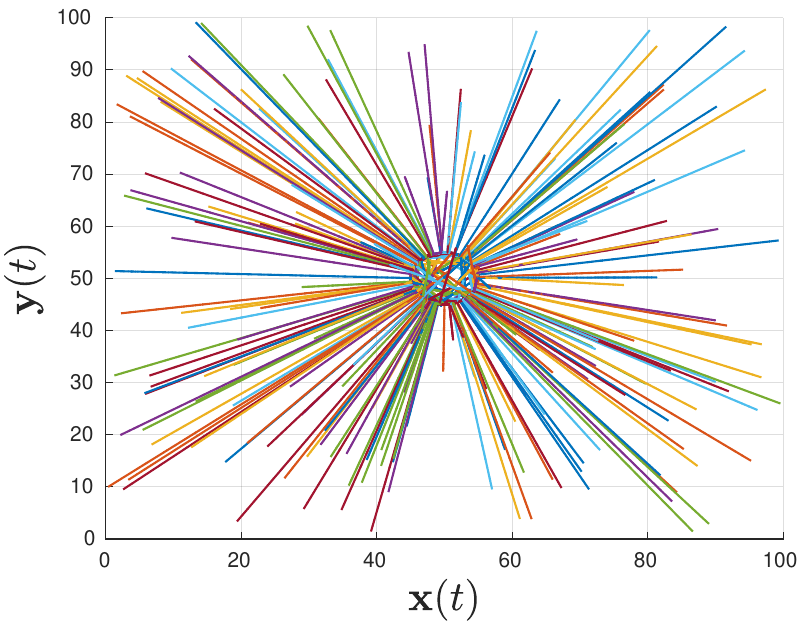}
         \subcaption{$x-t$ axis view at $j=1000$.}
         \label{fig:tf_xyview}
     \end{subfigure}%
        \caption{{Simulation of trajectory generation for 100 agents for the STL formula \eqref{eq:100agents}.}}
        \label{fig:100agents}
\end{figure*}

\begin{figure}
    \centering
    \includegraphics[width=0.8\columnwidth]{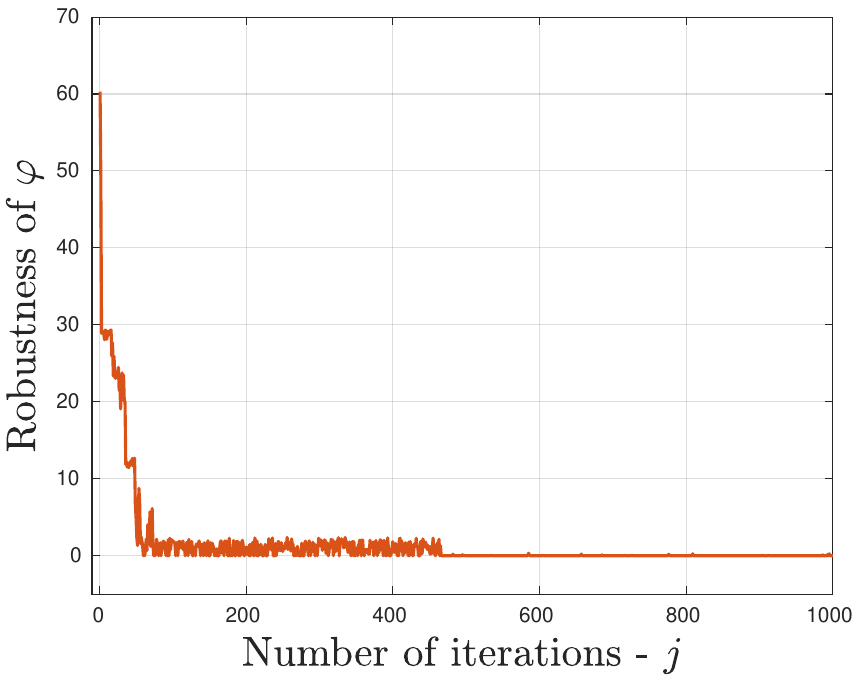}
    \caption{{Robustness of the STL formula in \eqref{eq:100agents}}.}
    \label{fig:robust}
\end{figure}

\subsubsection{Overall case study} In this case study, we demonstrate the application of the aforementioned scenarios by setting up the following tasks:

\begin{itemize}
    \item Agent 1 always stays above 8 units.
    \item Agents 2 and 4 are required to satisfy the predicate $x_2^2+x_4^2\leq 2$ within the time interval $[10,30]$[s].
    \item Agent 3 is required to track an exponential path within the time interval $[20,60]$[s].
    \item Agent 2 is required to repeatedly visit Agent 1 and Agent 3 every $10$s within the interval $[30,50]$[s].
    \item Agent 1 is required to maintain at least 1 unit distance from the other three agents within the interval $[80,100]$[s].
\end{itemize}
The STL formula for the above tasks is as follows: 
\begin{align*}
    \varphi&\quad  = (x_1\geq 8)\ \land\ \mathcal{G}_{[10,30]}(x_2^2+x_4^2\leq 2)\land \\
     &\mathcal{G}_{[20,60]}(\|x_3-50\exp(-0.1t)\|\leq 0.05) \land \\
    &  \mathcal{G}_{[30,50]}\mathcal{F}_{[0,10]} \Big((\|x_2-x_1\|\leq 0.5)\land(\|x_2-x_3\|\leq 0.5)\Big)\land\\
    & \mathcal{F}_{[79.9,80.1]}\mathcal{G}_{[0,20]} \Big((\|x_1-x_2\|\geq 1)\land (\|x_1-x_3\|\geq 1) \\
    &\land (\|x_1-x_4\|\geq 1)\Big)
\end{align*}
The parameter $L$ was increased to 1000, and $\eta$ was decreased to 0.001.
In Figure \ref{fig:sim:overall}, we show the resulting trajectories of each agent generated by \texttt{MAPS}$^2$ satisfying the above STL formula. The maximum computation time by any agent is $4.611$[s].

%\begin{rmrk}
%{To guarantee completeness, our focus in this work is directed towards the planning problem, specifically the generation of trajectories that fulfil a specific criterion, rather than the mechanics of how the agent moves or the precise control techniques used to execute the trajectory. This approach leads to the production of non-smooth trajectories, as seen in the simulations. To address this, we can apply a smoothing procedure to the trajectories using B-Splines, taking into account the velocity and acceleration constraints of the robots, see \cite{lapandic2023kinodynamic}. Furthermore, to the best of our knowledge, there has been no prior study that tackles the distributed multi-robot STL planning problem under nonlinear, nonconvex coupled constraints; thus a comparison study is not in order.}
%\end{rmrk}

%\begin{rmrk}
%While the computational advantages of a distributed algorithm may be limited or absent when dealing with even large number of robots, such as in the thousands, due to the existence of  efficient commercial  optimisers, the added advantage of redundancy within a multi-robot system is still crucial. This outweighs the benefits of centralised implementation.
%\end{rmrk}

\begin{figure*}
     \centering
     \begin{subfigure}[t]{0.325\textwidth}
         \centering
         \includegraphics[width=\textwidth]{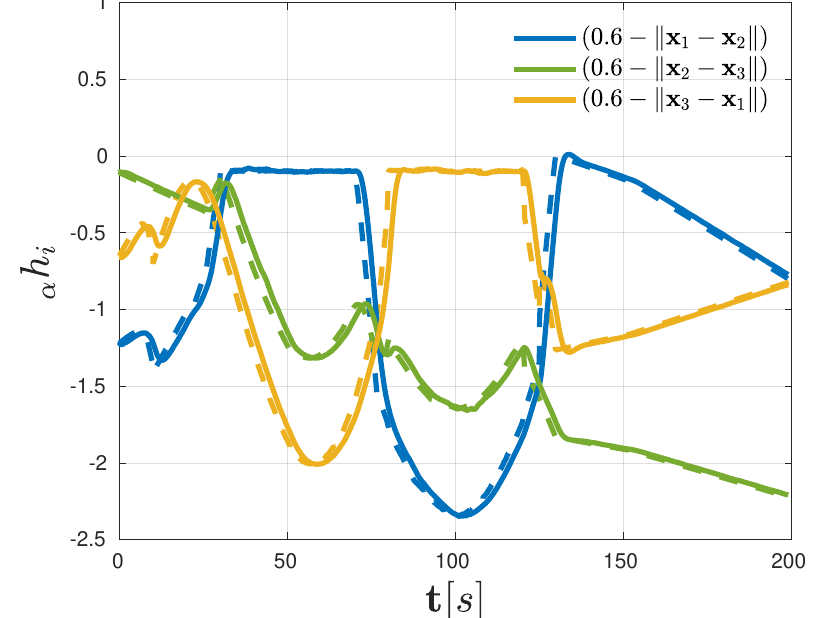}
         \subcaption{Collision constraints.}
         \label{fig:collision_all}
     \end{subfigure}%
     \begin{subfigure}[t]{0.325\textwidth}
         \centering
         \includegraphics[width=\textwidth]{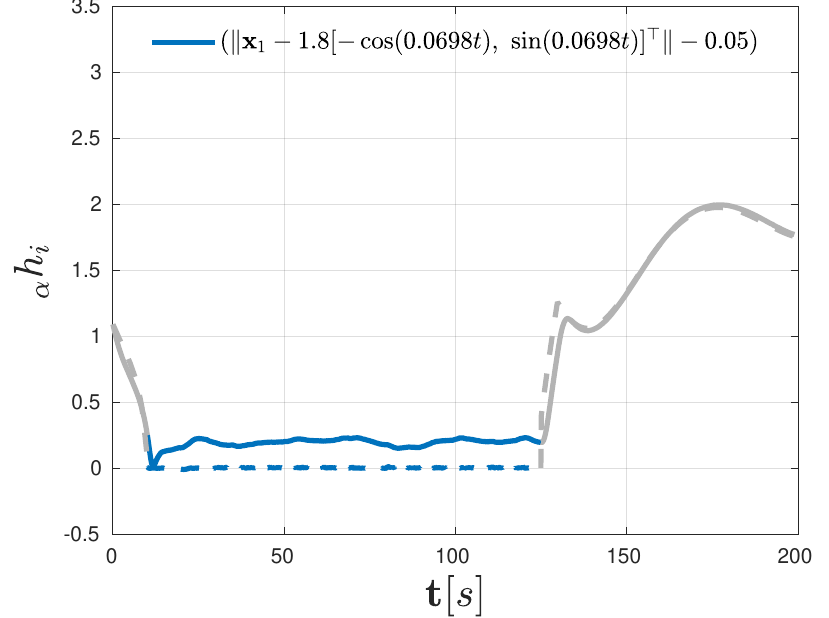}
         \subcaption{Base 1 trajectory tracking.}
         \label{fig:base_circle}
     \end{subfigure}%
     \begin{subfigure}[t]{0.325\textwidth}
         \centering
         \includegraphics[width=\textwidth]{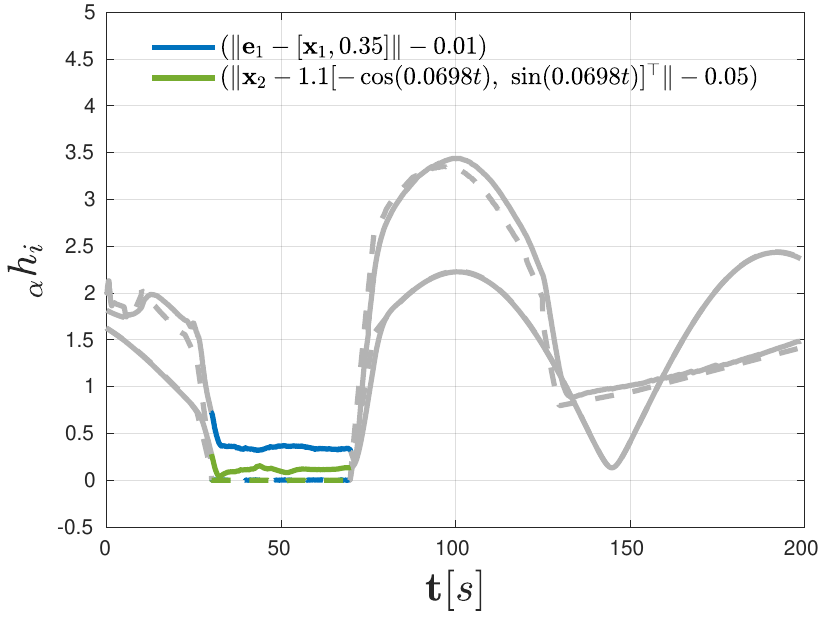}
         \subcaption{Virtual point tracker by base 2 and end effector 1.}
         \label{fig:base_ee1}
     \end{subfigure}\\
     \begin{subfigure}[t]{0.325\textwidth}
         \centering
         \includegraphics[width=\textwidth]{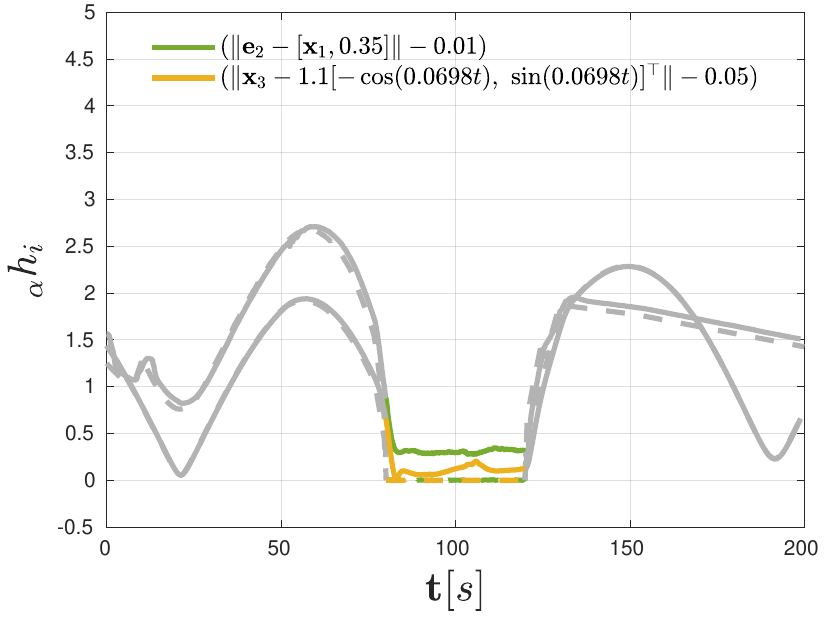}
         \subcaption{Virtual point tracker by base 3 and end effector 2.}
         \label{fig:base_ee2}
     \end{subfigure}%
     \begin{subfigure}[t]{0.325\textwidth}
         \centering
         \includegraphics[width=\textwidth]{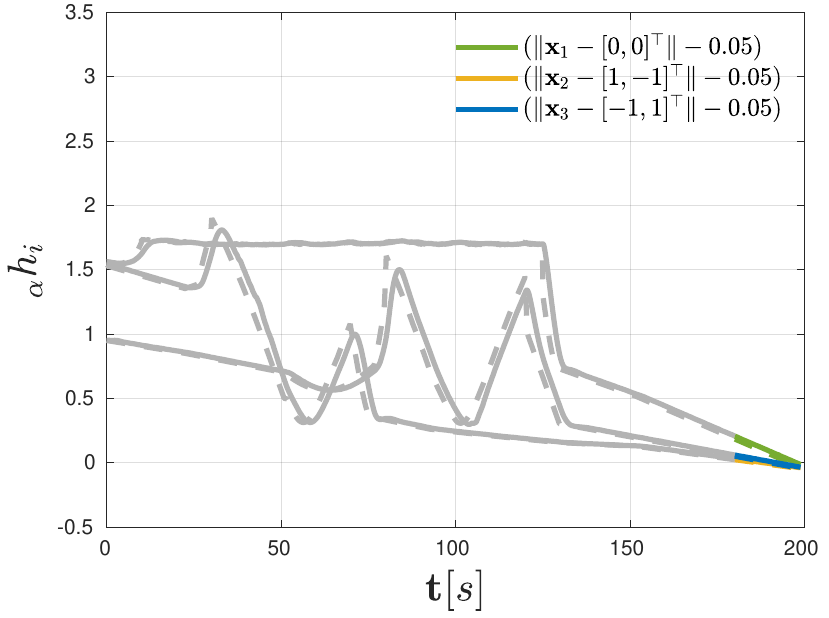}
         \subcaption{Desired final position of bases.}
         \label{fig:final1}
     \end{subfigure}%
     \begin{subfigure}[t]{0.325\textwidth}
         \centering
         \includegraphics[width=\textwidth]{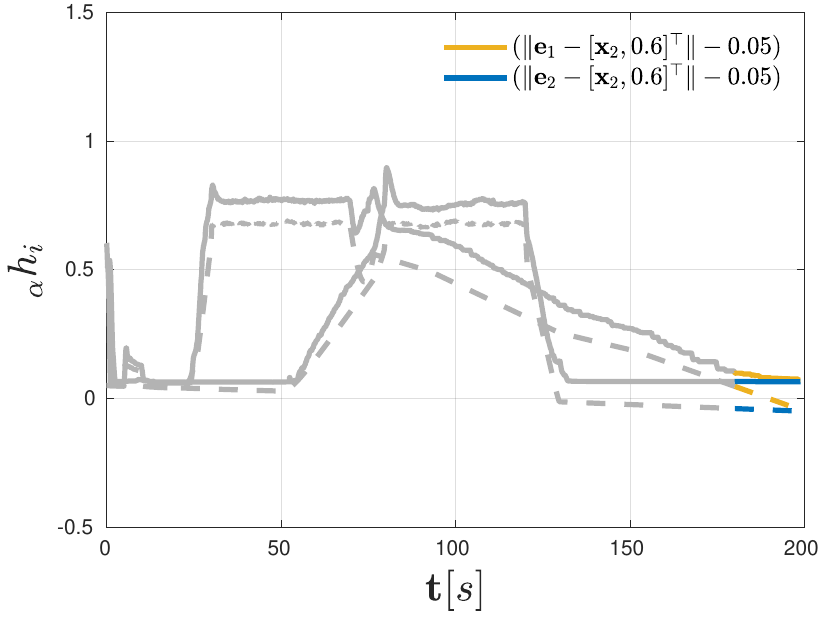}
         \subcaption{Desired final position of end effectors.}
         \label{fig:final2}
     \end{subfigure}%
        \caption{Experimental verification of MAPS$^2$ with the setup in Figure \ref{fig:setup}.}
        \label{fig:experiment_plots}
        \vspace{-0.4cm}
\end{figure*}

\section{Experiments}\label{sec:experiments}
We now present an experimental demonstration of the proposed algorithm. The multi-robot setup involves three robots, as shown in Figure \ref{fig:setup}, and consists of 3 mobile bases and two 6-DOF manipulator arms. The locations of the three bases are denoted as $\mathbf{x}_1\in\mathbb{R}^2$, $\mathbf{x}_2\in\mathbb{R}^2$, and $\mathbf{x}_3\in\mathbb{R}^2$, respectively. Base 2 and base 3 are equipped with manipulator arms, whose end-effector positions are represented as $\mathbf{e}_1\in\mathbb{R}^3$ and $\mathbf{e}_2\in\mathbb{R}^3$, respectively.

The STL formula defining the tasks is the following,
\begin{align*}
    &\varphi =  \|\mathbf{x}_1-\mathbf{x}_2\|\geq 0.6 \land
             \|\mathbf{x}_2-\mathbf{x}_3\|\geq 0.6 \land
             \|\mathbf{x}_3-\mathbf{x}_1\|\geq 0.6 \land\\
            & \mathcal{G}_{[10,125]} \|\mathbf{x}_1-1.8[-\cos{0.0698t},\sin(0.0698t)]^\top\|\leq 0.05 \land \\
            & \mathcal{G}_{[30,70]} \|\mathbf{e}_1-[\mathbf{x}_1^\top,0.35]^\top\|\leq 0.01 \land \\
            & \mathcal{G}_{[30,70]} \|\mathbf{x}_2-1.1[-\cos{0.0698t},\sin(0.0698t)]^\top\|\leq 0.05 \land \\
            & \mathcal{G}_{[80,120]} \|\mathbf{e}_2-[\mathbf{x}_1^\top,0.35]^\top\|\leq 0.01 \land \\
            & \mathcal{G}_{[80,120]} \|\mathbf{x}_3-1.1[-\cos{0.0698t},\sin(0.0698t)]^\top\|\leq 0.05 \land \\
            & \mathcal{F}_{[180,200]} \|\mathbf{x}_1-[0,0]^\top\|\leq 0.05 \land \\
            & \mathcal{F}_{[180,200]} \Big(\|\mathbf{x}_2-[1,-1]\|\leq 0.05 \land \|\mathbf{e}_1-[\mathbf{x}_2, 0.6]\|\leq 0.05 \Big)\land\\
            & \mathcal{F}_{[180,200]} \Big(\|\mathbf{x}_3-[-1,1]\|\leq 0.05 \land \|\mathbf{e}_2-[\mathbf{x}_3, 0.6]\|\leq 0.05 \Big).
\end{align*}

The above task involves collision avoidance constraints that are always active given by the subformula $\bar{\varphi}_1=(\|\mathbf{x}_1-\mathbf{x}_2\|\geq 0.6)\land(\|\mathbf{x}_2-\mathbf{x}_3\|\geq 0.6)\land (\|\mathbf{x}_3-\mathbf{x}_1\|\geq 0.6)$. Next, in the duration $[10,125]$[s], base 1 surveils the arena and follows a circular time varying trajectory given by the subformula $\bar{\varphi}_2=(\mathcal{G}_{[10,125]} \|\mathbf{x}_1-c_1(t)\|\leq 0.05)$ where $c_1(t)$ is the circular trajectory. In the duration $[30,70]$[s], end-effector 1 tracks a virtual point $0.35$[m] over base 1 to simulate a pick-and-place task, given by the subformula $\bar{\varphi}_3=\mathcal{G}_{[30,70]} \|\mathbf{e}_1-[\mathbf{x}_1^\top,0.35]^\top\|\leq 0.01 \land \mathcal{G}_{[30,70]} \|\mathbf{x}_2-c_2(t)\|\leq 0.05$ where $c_2(t)$ is the circular trajectory. Similarly, in the duration $[80,120]$[s], end-effector 2 takes over the task to track a virtual point $0.35$[m] over base 1, given by the subformula $\bar{\varphi}_4=\mathcal{G}_{[80,120]} \|\mathbf{e}_2-[\mathbf{x}_1^\top,0.35]^\top\|\leq 0.01 \land \mathcal{G}_{[80,120]} \|\mathbf{x}_3-c_2(t)\|\leq 0.05$. Finally, eventually in the duration $[180,200]$[s], the robots assume a final position given by the subformula $\bar{\varphi_5}=\mathcal{F}_{[180,200]} \|\mathbf{x}_1-[0,0]^\top\|\leq 0.05 \land  \mathcal{F}_{[180,200]} \Big(\|\mathbf{x}_2-[1,-1]\|\leq 0.05 \land \|\mathbf{e}_1-[\mathbf{x}_2, 0.6]\|\leq 0.05 \Big)\land\mathcal{F}_{[180,200]} \Big(\|\mathbf{x}_3-[-1,1]\|\leq 0.05 \land \|\mathbf{e}_2-[\mathbf{x}_3, 0.6]\|\leq 0.05 \Big).$

The results are shown in Figure \ref{fig:experiment_plots}, where the x-axis represents time in seconds, and the y-axis represents the predicate functions defined by \eqref{state_constraints}. The dashed line in the plots represents the predicate functions of the trajectories obtained by solving the optimisation problem \eqref{tro:eq:optimisation}, while the solid line represents the predicate functions of the actual trajectories by the robots. In the context of \eqref{state_constraints}, negative values indicate task satisfaction. However, due to the lack of an accurate model of the robots and the fact that the optimisation solution converges to the boundary of the constraints, the tracking is imperfect, and we observe slight violations of the formula by the robots in certain cases. Nonetheless, the trajectories generated by the algorithm do not violate the STL formula. The coloured lines represent the functions that lie within the validity domain of the formula.
Figure \ref{fig:collision_all} shows that the collision constraint imposed on all 3 bases is not violated, and they maintain a separation of at least 60 cm. In Figure \ref{fig:base_circle}, base 1 tracks a circular trajectory in the interval $[10, 125]$ seconds. In Figures \ref{fig:base_ee1} and \ref{fig:base_ee2}, the end effectors mounted on top of bases 2 and 3 track a virtual point over the moving base 1 sequentially. In the last 20 seconds, the bases and end effectors move to their desired final positions, as seen in Figures \ref{fig:final1} and \ref{fig:final2}. The maximum computation time by any robot is $3.611$[s]. Figure \ref{fig:snapshots} shows front-view and side-view at different time instances during the experimental run\endnote{The video of the experiments can be found here: \url{https://youtu.be/YkuiPuOerMg}}. 

\begin{figure*}
     \centering
     \begin{subfigure}[t]{0.49\textwidth}
         \centering
         \includegraphics[width=\textwidth]{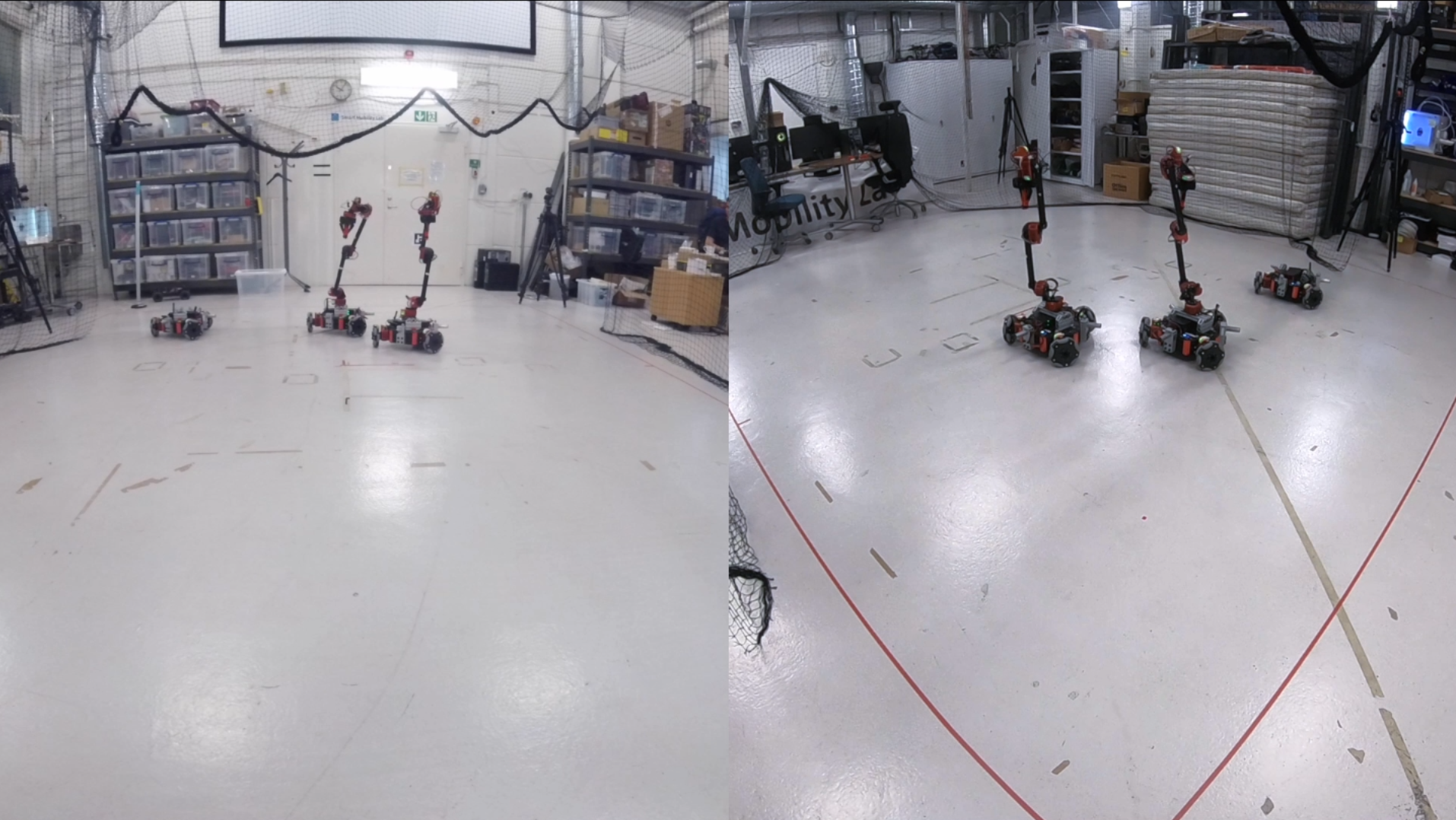}
         \subcaption{At $t=12$[s], when $\bar{\varphi}_1$ and $\bar{\varphi}_2$ are active.}
         %\label{fig:collision_all}
     \end{subfigure}
     \begin{subfigure}[t]{0.49\textwidth}
         \centering
         \includegraphics[width=\textwidth]{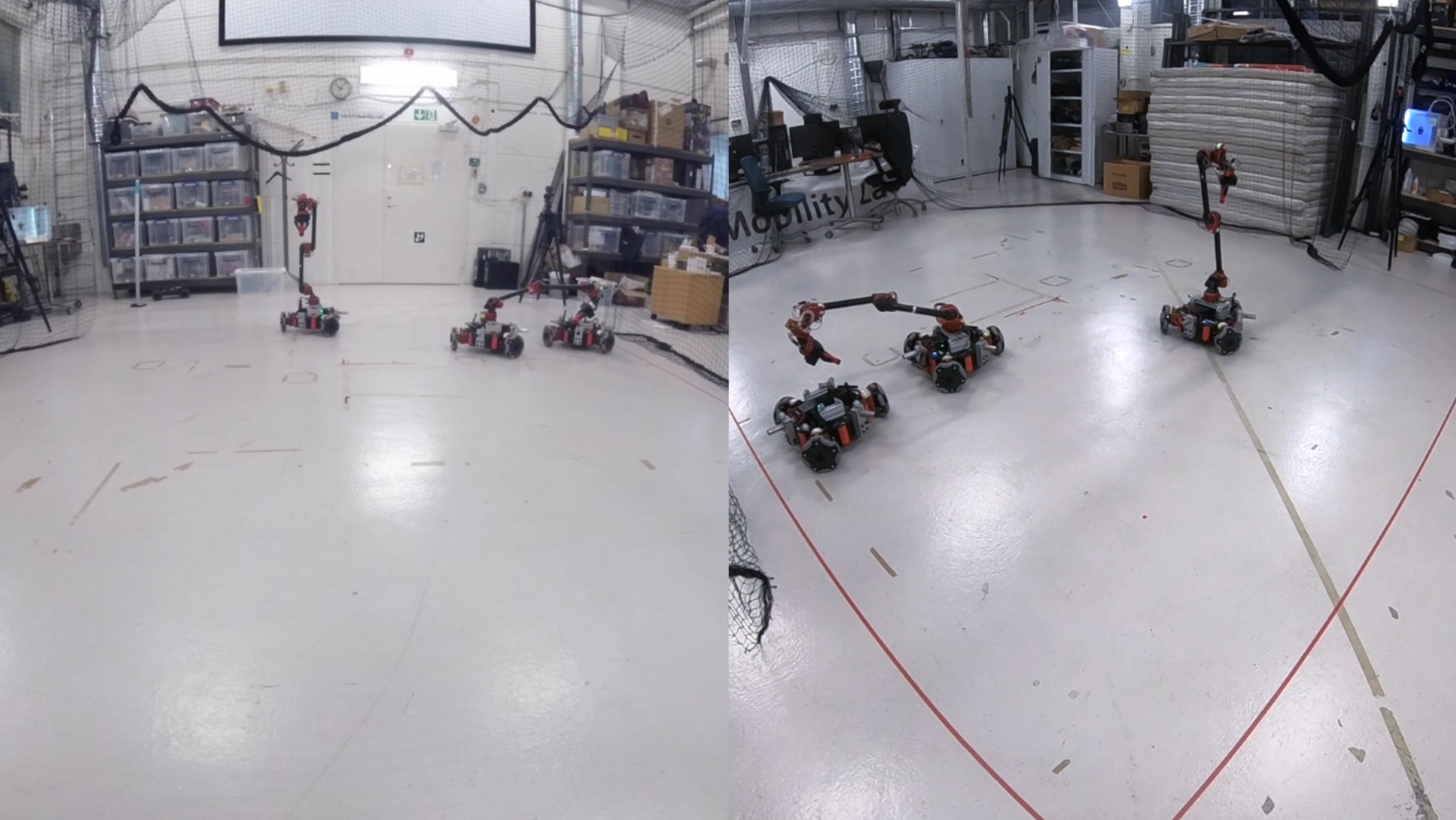}
         \subcaption{At $t=50$[s], when $\bar{\varphi}_1$, $\bar{\varphi}_2$ and $\bar{\varphi}_3$ are active.}
         %\label{fig:base_circle}
     \end{subfigure}\\
     \begin{subfigure}[t]{0.49\textwidth}
         \centering
         \includegraphics[width=\textwidth]{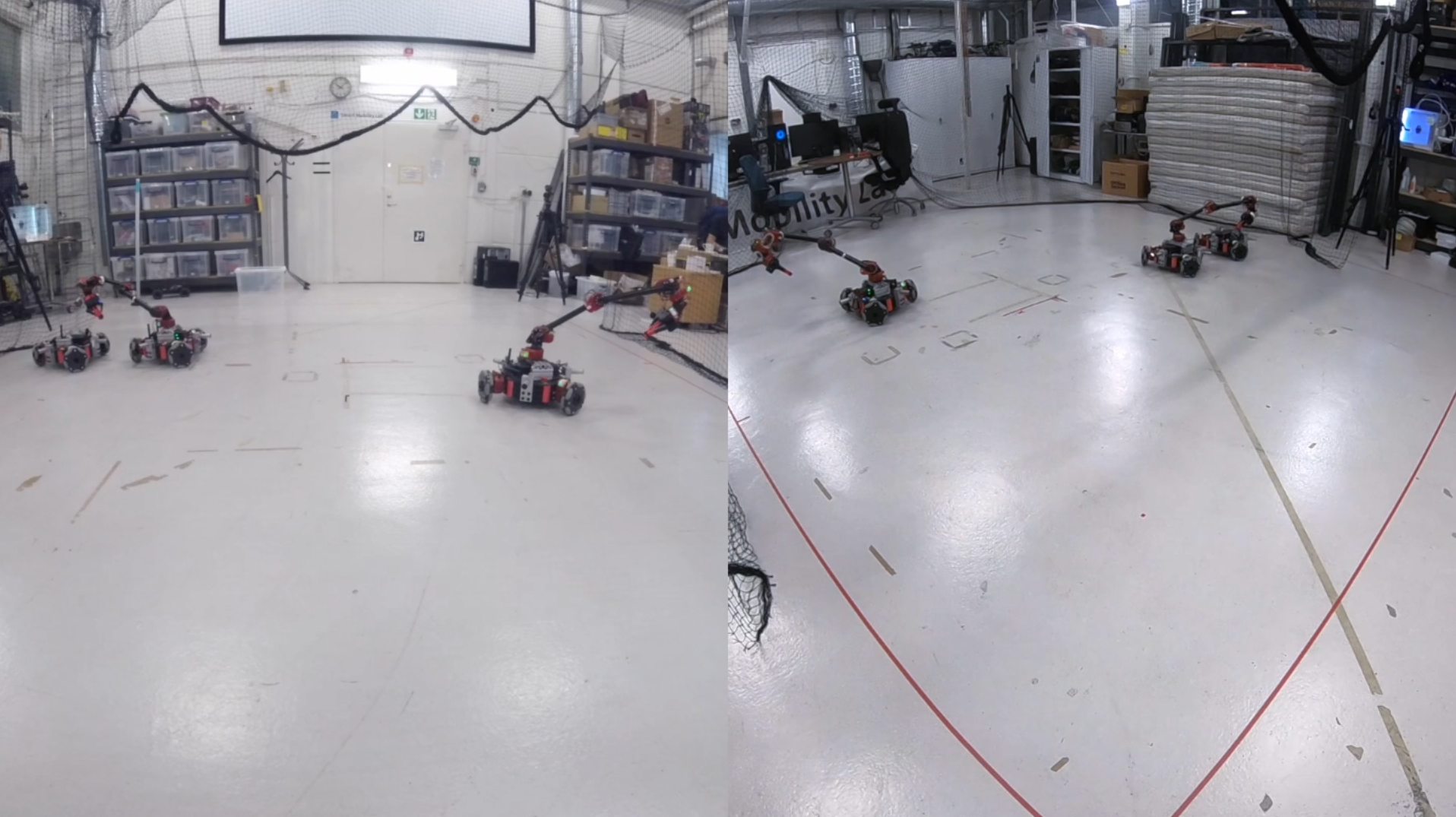}
         \subcaption{At $t=100$[s], when $\bar{\varphi}_1$, $\bar{\varphi}_2$ and $\bar{\varphi}_4$ are active.}
         %\label{fig:base_ee1}
     \end{subfigure}
     \begin{subfigure}[t]{0.49\textwidth}
         \centering
         \includegraphics[width=\textwidth]{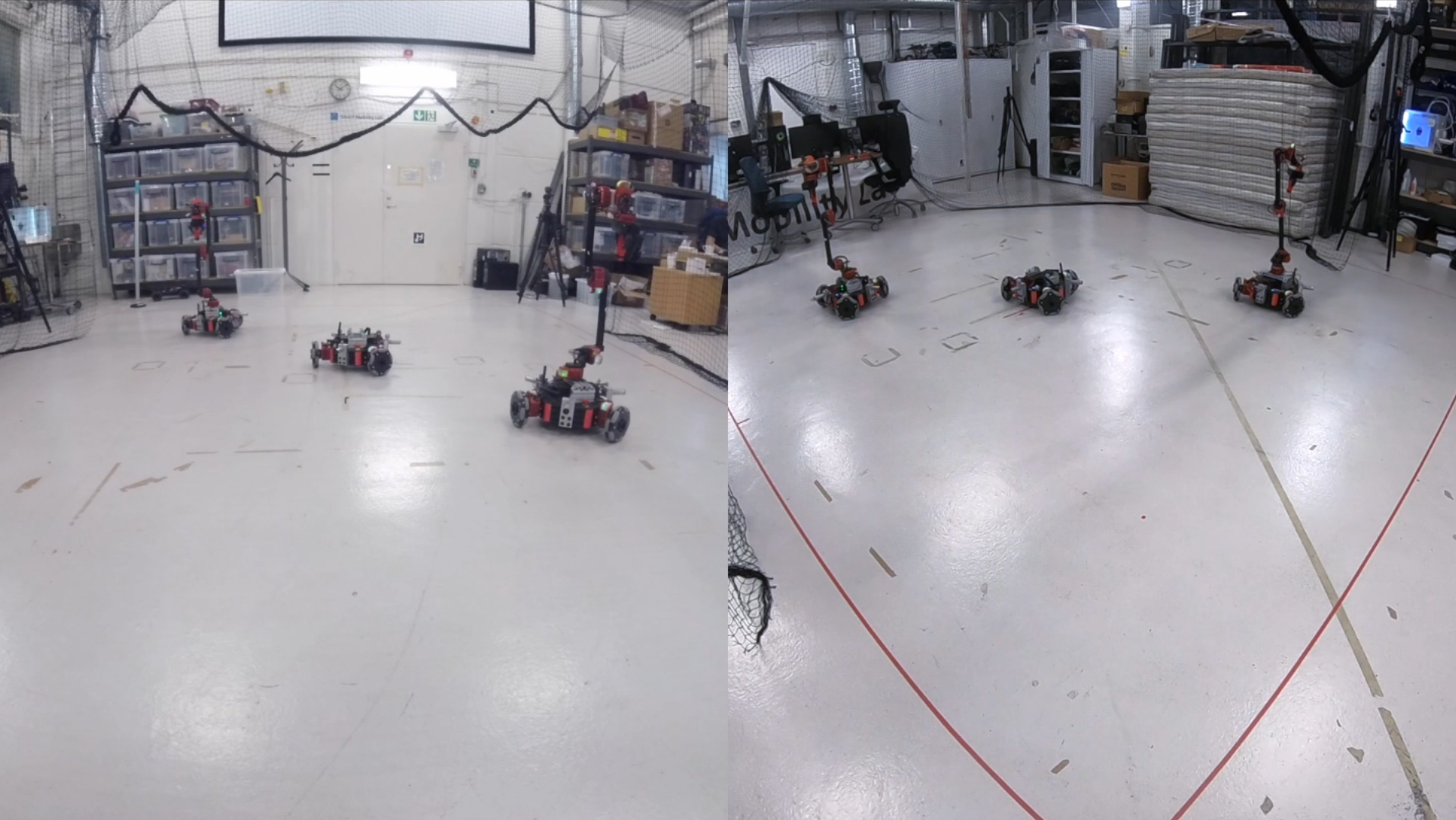}
         \subcaption{At $t=190$[s], when $\bar{\varphi}_1$ and $\bar{\varphi}_5$ are active.}
         %\label{fig:base_ee2}
     \end{subfigure}%
        \caption{Front-view and side-view during experimental run with the setup in Figure \ref{fig:setup}.}
        \label{fig:snapshots}
        \vspace{-0.4cm}
\end{figure*}
%%%%%%%%%%%%%%%%%%%%%%%%%%%%%%%%%%%%%%%%%%%%%%%%%%%%%%%%%%%%%%%%%%%%%%%%%%%%%%%%

\section{Conclusion}\label{sec:conclusion}

This work proposed MAPS$^2$, a distributed planner that solves the multi-robot motion-planning problem subject to tasks encoded as STL constraints. By using the notion of validity domain and formulating the optimisation problem as shown in \eqref{tro:eq:optimisation}, MAPS$^2$ transforms the spatio-temporal problem into a spatial planning task, for which efficient optimisation algorithms already exist. Task satisfaction is probabilistically guaranteed in a distributed manner by presenting an optimisation problem that necessitates communication only between robots that share coupled constraints. Extensive simulations involving benchmark formulas and experiments involving varied tasks highlight the algorithms functionality. { Future work involves incorporating dynamical constraints such as velocity and acceleration limits into the optimisation problem. }
%%%%%%%%%%%%%%%%%%%%%%%%%%%%%%%%%%%%%%%%%%%%%%%%%%%%%%%%%%%%%%%%%%%%%%%%%%%%%%%%

%%%%%%%%%%%%%%%%%%%%%%%%%%%%%%%%%%%%%%%%%%%%%%%%%%%%%%%%%%%%%%%%%%%%%%%%%%%%%%%%

%%%%%%%%%%%%%%%%%%%%%%%%%%%%%%%%%%%%%%%%%%%%%%%%%%%%%%%%%%%%%%%%%%%%%%%%%%%%%%%%

\begin{acks}
This work was supported by the ERC CoG LEAFHOUND, the Swedish Research Council (VR), the Knut och Alice Wallenberg Foundation (KAW) and the H2020 European Project CANOPIES.
\end{acks}

\theendnotes

\bibliography{IJRR_final_submission/references_final}
\bibliographystyle{apalike}

\end{document}